\documentclass[lettersize,journal]{IEEEtran}
\usepackage{amsmath,amsfonts}
\usepackage{algorithmic}
\usepackage{algorithm}
\usepackage{array}
\usepackage[caption=false,font=normalsize,labelfont=sf,textfont=sf]{subfig}
\usepackage{url}
\usepackage{textcomp}
\usepackage{stfloats}
\usepackage{url}
\usepackage{verbatim}
\usepackage{graphicx}
\usepackage{cite}
\usepackage{amsmath} 
\usepackage{multirow}   
\usepackage{booktabs}   
\usepackage{xcolor}     
\usepackage{graphicx}   

\usepackage{mathtools}

\usepackage{hyperref}
\hypersetup{hidelinks}

\usepackage{newtxtext} 
\usepackage{newtxmath} 

\begin{document}

\title{Building Extraction from Remote Sensing Imagery under Hazy and Low-light Conditions: Benchmark and Baseline}

\author{Feifei Sang$^\dagger$,
        Wei Lu$^\dagger$,
        Hongruixuan Chen,
        Sibao Chen$^{*}$,
        and Bin Luo
        
\thanks{This work was supported in part by the NSFC Key Project of Joint Fund for Enterprise Innovation and Development under Grant U24A20342, and in part by the National Natural Science Foundation of China under Grant 62576006 and 61976004. ($^{\dagger}$ Equal contribution, $^{*}$ Corresponding author: Sibao Chen.)}%
\thanks{Feifei Sang, Wei Lu, Sibao Chen, and Bin Luo are with the MOE Key Lab of ICSP, Anhui Provincial Key Lab of Multimodal Cognitive Computation, IMIS Lab of Anhui Province, School of Computer Science and Technology, Anhui University, Hefei, China (e-mail: ffeisang@163.com; luwei\_ahu@qq.com; sbchen@ahu.edu.cn; luobin@ahu.edu.cn).}%
\thanks{Hongruixuan Chen is with the Graduate School of Frontier Sciences, The University of Tokyo, Chiba, 277-8561, Japan (e-mail: qschrx@gmail.com).}%
}
\markboth{Journal of \LaTeX\ Class Files,~Vol.~18, No.~9, September~2020}%
{Building Extraction from Remote Sensing Imagery under Hazy and Low-light Conditions: Benchmark and Baseline}

\maketitle
\begin{abstract}
Building extraction from optical Remote Sensing (RS) imagery suffers from performance degradation under real-world hazy and low-light conditions. However, existing optical methods and benchmarks focus primarily on ideal clear-weather conditions. While SAR offers all-weather sensing, its side-looking geometry causes geometric distortions. To address these challenges, we introduce HaLoBuilding, the first optical benchmark specifically designed for building extraction under hazy and low-light conditions. By leveraging a same-scene multitemporal pairing strategy, we ensure pixel-level label alignment and high fidelity even under extreme degradation. Building upon this benchmark, we propose HaLoBuild-Net, a novel end-to-end framework for building extraction in adverse RS scenarios. At its core, we develop a Spatial-Frequency Focus Module (SFFM) to effectively mitigate meteorological interference on building features by coupling large receptive field attention with frequency-aware channel reweighting guided by stable low-frequency anchors. Additionally, a Global Multi-scale Guidance Module (GMGM) provides global semantic constraints to anchor building topologies, while a Mutual-Guided Fusion Module (MGFM) implements bidirectional semantic-spatial calibration to suppress shallow noise and sharpen weather-induced blurred boundaries. Extensive experiments demonstrate that HaLoBuild-Net significantly outperforms state-of-the-art methods and conventional cascaded restoration-segmentation paradigms on the HaLoBuilding dataset, while maintaining robust generalization on WHU, INRIA, and LoveDA datasets. The source code and datasets are publicly available at: \url{https://github.com/AeroVILab-AHU/HaLoBuilding}.
\end{abstract}

\begin{IEEEkeywords}
Building extraction, hazy and low-light imagery, benchmark dataset.
\end{IEEEkeywords}
\section{Introduction}
\IEEEPARstart{B}{uildings} constitute the core component of urban infrastructure, supporting the majority of human activities. Accurate building extraction from high-resolution Remote Sensing (RS) imagery not only provides essential data support for smart city construction and urban planning~\cite{ 8421076}, but also plays a critical role in emergency response to sudden disasters such as earthquakes and explosions~\cite{zheng2021building}.

Early building extraction relied on hand-crafted features with traditional classifiers. With the rise of deep learning, neural networks have dominated various vision tasks~\cite{LU2026431, Lu_2025_ICCV, 11313649} and been successfully applied to building extraction. Convolutional Neural Networks (CNNs)~\cite{ZHANG2020111912} and Fully Convolutional Networks (FCNs)~\cite{7478072} improved hierarchical feature learning and achieved high accuracy, but their locality limits modeling of large-scale variations. To address this, Vision Transformers were introduced to capture long-range dependencies\cite{chen2021building}. Hybrid CNN–Transformer models~\cite{2022Transformer}, especially those based on Swin Transformer, enhance performance by combining global context with local details, while Easy-Net~\cite{10376153} further balances accuracy and efficiency through lightweight design and effective fusion.

\begin{figure}[t]
	\centering 
	\includegraphics[width=0.48\textwidth]{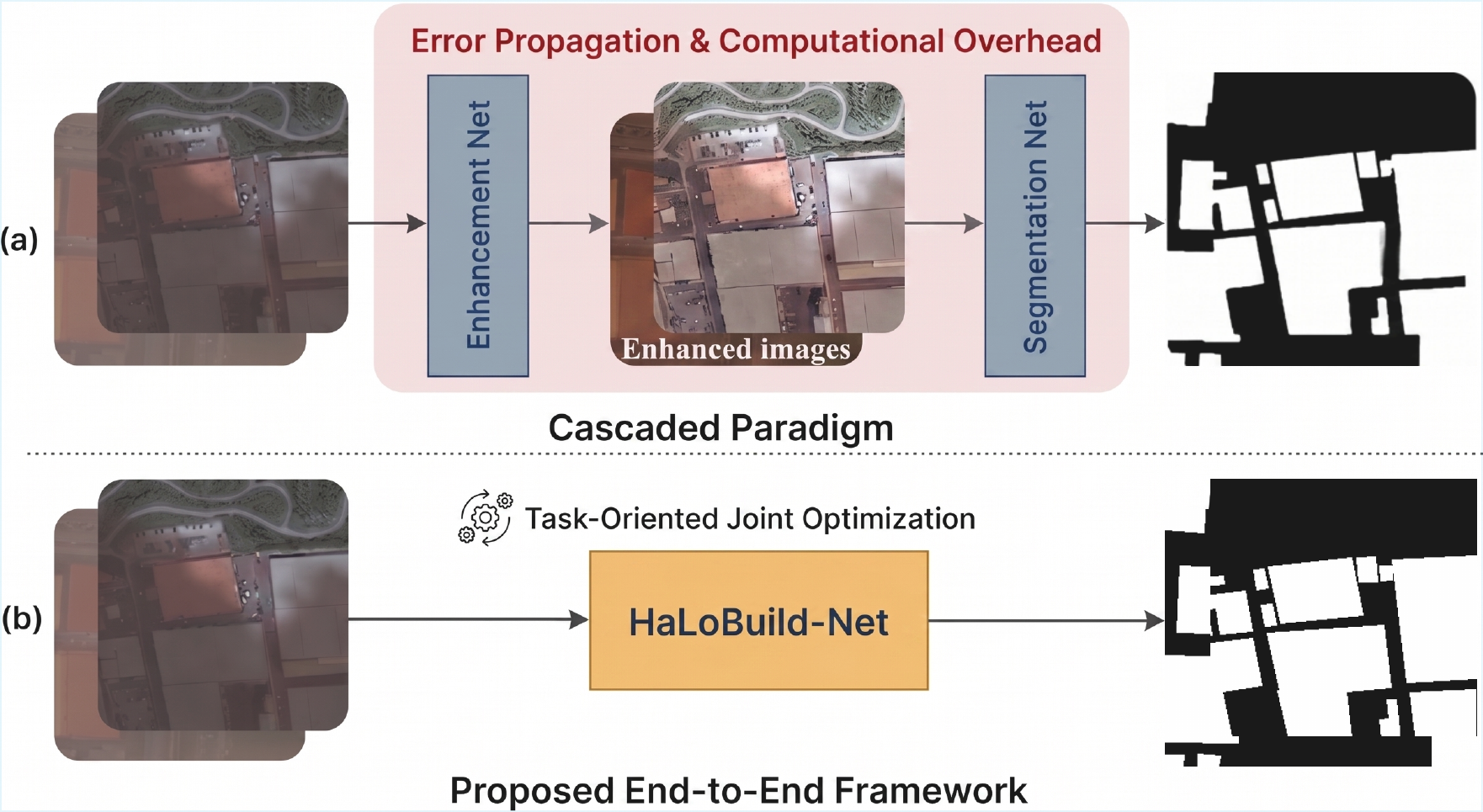}\vspace{-2mm}
	\caption{A conceptual comparison of building extraction paradigms. (a) The cascaded paradigm suffers from inherent error propagation, artifacts, and computational overhead due to its disjointed stages. (b) Our end-to-end HaLoBuild-Net bypasses explicit enhancement via task-oriented joint optimization for robust high-fidelity extraction.}
	\label{zhutu}
\end{figure}

Nevertheless, signal-to-noise ratio reduction and texture loss caused by adverse weather frequently lead to the failure of attention mechanisms in existing models, triggering severe missed and false detections. Early cascaded strategies following an enhancement-first and segmentation-later paradigm suffer from computational redundancy and inconsistent optimization objectives. As conceptually compared in Fig.~\ref{zhutu}(a), this disjointed pipeline introduces significant computational overhead and inevitably propagates enhancement-induced artifacts into the segmentation head, thereby exacerbating performance degradation. To overcome these bottlenecks, transitioning towards a task-oriented end-to-end framework (Fig.~\ref{zhutu}(b)) that bypasses explicit image restoration is highly desirable. Although recent end-to-end networks~\cite{10611148} attempt joint optimization, they still depend on auxiliary enhancement sub-networks. More recently, frequency-domain adaptation methods such as NightAdapter~\cite{11094333} achieve robust learning directly in feature space by decoupling illumination interference, realizing nighttime segmentation without relying on explicit image enhancement. However, these methods are primarily designed for street-view scenes and rely heavily on side textures and light sources. This makes it difficult to recover fine edges of dense buildings in top-down RS perspectives.

While Synthetic Aperture Radar (SAR) provides a weather-independent alternative, its side-looking geometry introduces inherent and severe geometric distortions, such as layover and shadowing. These artifacts lead to a fundamental displacement of building footprints, fundamentally compromising the geometric precision required for building footprint extraction. Conversely, optical imagery, even when severely degraded, preserves two irreplaceable properties. First, it maintains pristine nadir-view geometric fidelity. Although visual cues are obscured, the pixels corresponding to a building's location remain spatially accurate. The challenge is therefore a signal restoration problem, not a geometric correction one, which is often intractable for SAR. Second is the potential for semantic recovery. While features like roof boundaries are attenuated, their underlying information is latent, not absent. A deep network trained on appropriate data can learn a mapping function to recover discriminative building features from degraded observations, a form of semantic recovery that is fundamentally infeasible for radar signals due to their distinct physical measurement principles. Given that most operational building extraction methods are deeply reliant on optical sensors, developing robust algorithms to master degraded optical data presents a more practical path than switching to a geometrically flawed and semantically incompatible SAR sensor and developing entirely new extraction paradigms for it. However, a dedicated, large-scale optical benchmark for high-precision building extraction under hazy and low-light conditions is still lacking.

To fill this gap, we establish HaLoBuilding, the first large-scale benchmark specifically designed for building extraction under hazy and low-light conditions. To resolve the formidable challenges of data scarcity and degraded imagery annotation, we utilize a same-scene multitemporal pairing strategy to effectively transfer high-quality labels from clear reference images to their adverse-weather counterparts, followed by rigorous manual verification and refinement to ensure pixel-level alignment and high-fidelity annotations.

Based on this benchmark, we propose HaLoBuild-Net, a novel end-to-end building extraction network without explicit image enhancement. We design a Spatial-Frequency Focus Module (SFFM) to achieve dual-domain collaborative optimization by coupling spatial attention with a frequency-aware reweighting mechanism. The network further integrates a Global Multi-scale Guidance Module (GMGM) and a Mutual-Guided Fusion Module (MGFM). Specifically, the GMGM aggregates multi-scale shallow features to provide global semantic constraints, effectively anchoring building topologies even under severe degradation. Simultaneously, the MGFM performs bidirectional semantic-spatial calibration. It utilizes deep semantics as a spatial filter to mitigate meteorological noise in shallow layers, while also leveraging shallow details as structural anchors to compensate for boundary diffusion in the deep features.

The main contributions are summarized as follows:
\begin{enumerate}
    \item We introduce HaLoBuilding, the 
    first optical benchmark dataset specifically curated for building extraction in hazy and low-light RS scenarios. It provides high-fidelity, pixel-level aligned annotations through a multitemporal pairing and manual refinement strategy.
    \item We propose HaLoBuild-Net, a novel end-to-end framework that decouples complex environmental degradation from building features in a dual-domain space. It integrates the SFFM for frequency-guided feature recalibration, the GMGM for topological anchoring, and the the MGFM for bidirectional semantic-spatial calibration to ensure geometric precision under complex degradations.
    \item Experiments demonstrate that HaLoBuild-Net outperforms both state-of-the-art methods and cascaded paradigms on HaLoBuilding, while maintaining robust generalization on WHU, INRIA, and LoveDA.
\end{enumerate}

\section{Related Work}
\subsection{Building Extraction in General Scenarios}
With the development of remote sensing technology, building extraction has evolved from early handcrafted feature engineering~\cite{DORNAIKA2016130} to deep learning-dominated end-to-end paradigms. Early studies primarily relied on fully convolutional networks~\cite{7478072}, combined with post-processing techniques~\cite{8933116} to optimize results. 

To overcome the limitations of fixed receptive fields in CNNs and the large scale variations of buildings, researchers have explored various strategies. Some have focused on optimizing feature interaction and context aggregation~\cite{9296550,9750048}. Others have employed multi-path designs to enhance feature discrimination~\cite{9212557, 9171479}, or utilized spatial and channel attention to suppress background noise~\cite{pan2019building}. For challenges related to small targets and long-range dependencies, significant enhancements in feature discrimination have been achieved through methods like hybrid dilated convolutions, self-attention encoding, and grid attention gating~\cite{9423811}. Additionally, to tackle boundary blurring and feature entanglement in buildings, researchers further proposed edge optimization and multi-task collaborative strategies: CBR-Net~\cite{GUO2022240} and BOMSC-Net~\cite{9716137} employed coarse-to-fine boundary refinement and directional field features, respectively, to sharpen edges and mitigate occlusion issues; FD-Net~\cite{10155230} innovatively decoupled features into semantic and edge subspaces for supervised guided fusion; while LCS~\cite{9924239} introduced a line segment collaborative segmentation framework, leveraging the complementary advantages of vector extraction and semantic segmentation to improve accuracy. Similarly, UANet~\cite{10418227} introduced an uncertainty-aware framework to refine feature representation and reduce prediction ambiguity around complex building edges. More recently, EGAFNet~\cite{10819433} further addressed boundary blurring and scale variation by integrating edge-guided boundary enhancement with scale-aware adaptive multiscale feature fusion.

Although CNNs excel in local detail extraction, their global modeling capability remains limited. To this end, Vision Transformers have been introduced into remote sensing due to their ability to capture long-range dependencies: using backbones such as Swin Transformer or SegFormer to enhance multi-scale global modeling~\cite{chen2022, 9775207} has become a trend. Meanwhile, to balance local details and global semantics, pyramid self-attention and non-local modules embedded in CNNs~\cite{s20247241, 8950134}, or dual-branch hybrid architectures~\cite{9296550,wang2022building}, have emerged as mainstream choices. Following this hybrid paradigm, MSHFormer~\cite{10908214} synergized multiscale local perception with global transformer modeling to balance feature extraction, while incorporating a dedicated edge enhancement module to sharpen building boundaries. Easy-Net~\cite{10376153} further balanced this trade-off by combining lightweight CNN backbones with efficient Transformer-based fusion blocks. To advance task universality, RSBuilding~\cite{10623867} adopted a foundation model approach, leveraging a cross-attention decoder to unify building extraction and change detection across diverse remote sensing scenes. Furthermore, in the context of domain adaptation, Xu et al.\cite{10891046} leveraged a hierarchical transformer to capture multi-level global context and integrated attraction field maps to explicitly enforce geometric structural consistency. To further advance end-to-end vectorization, HiT~\cite{10401962} utilized a hierarchical transformer with a polygon head to directly predict serialized vertices, thereby capturing fine-grained geometric structures at both vertex and edge levels.

\subsection{Building Extraction under Adverse Weather}
For image degradation problems, solutions have evolved from multi stage cascading to end to end feature adaptation. Early mainstream approaches adopted an enhancement first and segmentation later cascaded strategy, using image enhancement~\cite{guo2020zero} or dehazing algorithms~\cite{8237773} as preprocessing to restore visual quality before feeding into segmentation networks. However, this inherently introduces significant computational redundancy and inconsistency between enhancement and segmentation objectives. To overcome these limitations, end-to-end joint training has gained prominence. Researchers have incorporated auxiliary tasks to achieve mutual benefits between enhancement and segmentation: SFNet-N~\cite{9784832} and SKF~\cite{wu2023learningsemanticawareknowledgeguidance} introduced semantic priors into enhancement networks, guiding low-light image restoration via semantic consistency; nevertheless, such methods typically require additional enhancement sub-networks, significantly increasing model parameters and inference burden, making them largely unsuitable for efficient processing of large-scale remote sensing data. In contrast, robust learning directly in feature space offers a more efficient strategy. FeatEnHancer~\cite{hashmi2023featenhancer} proposed multi-scale enhancement in feature space, using task-driven attention to directly optimize hierarchical features and avoid redundancy from explicit image recovery. DTP~\cite{10378015} and RUAS~\cite{9914672} decoupled images into reflectance and illumination components for robust parsing; DAI-Net~\cite{du2024boosting} explored zero-shot day-night domain adaptation. However, these methods were primarily developed for street-view scenes, heavily relying on side textures and light source cues, thus struggling to effectively recover edge details of dense small buildings in remote sensing top-view perspectives.

\begin{figure} 
	\centering
	\includegraphics[width=\linewidth]{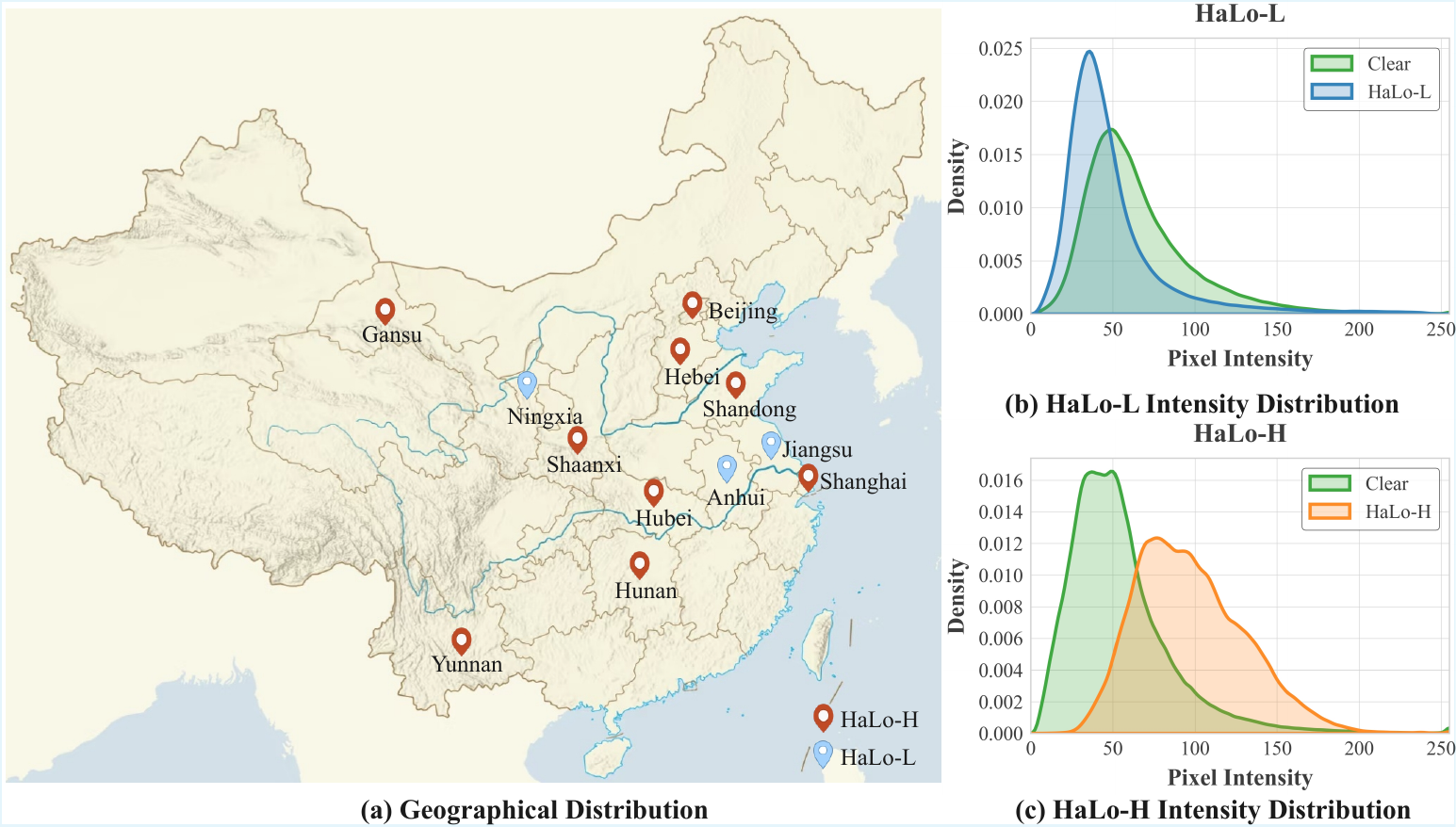}\vspace{-2mm}
	\caption{Geographical distribution and statistical characteristics of HaLoBuilding dataset. (a) Map illustrating sampling locations across multiple provinces to ensure geographical diversity. (b)(c) KDE analysis quantifying the profound domain shift between degraded and clear imagery. }
	\label{location}
\end{figure}

\begin{figure} 
	\centering
	\includegraphics[width=1\linewidth]{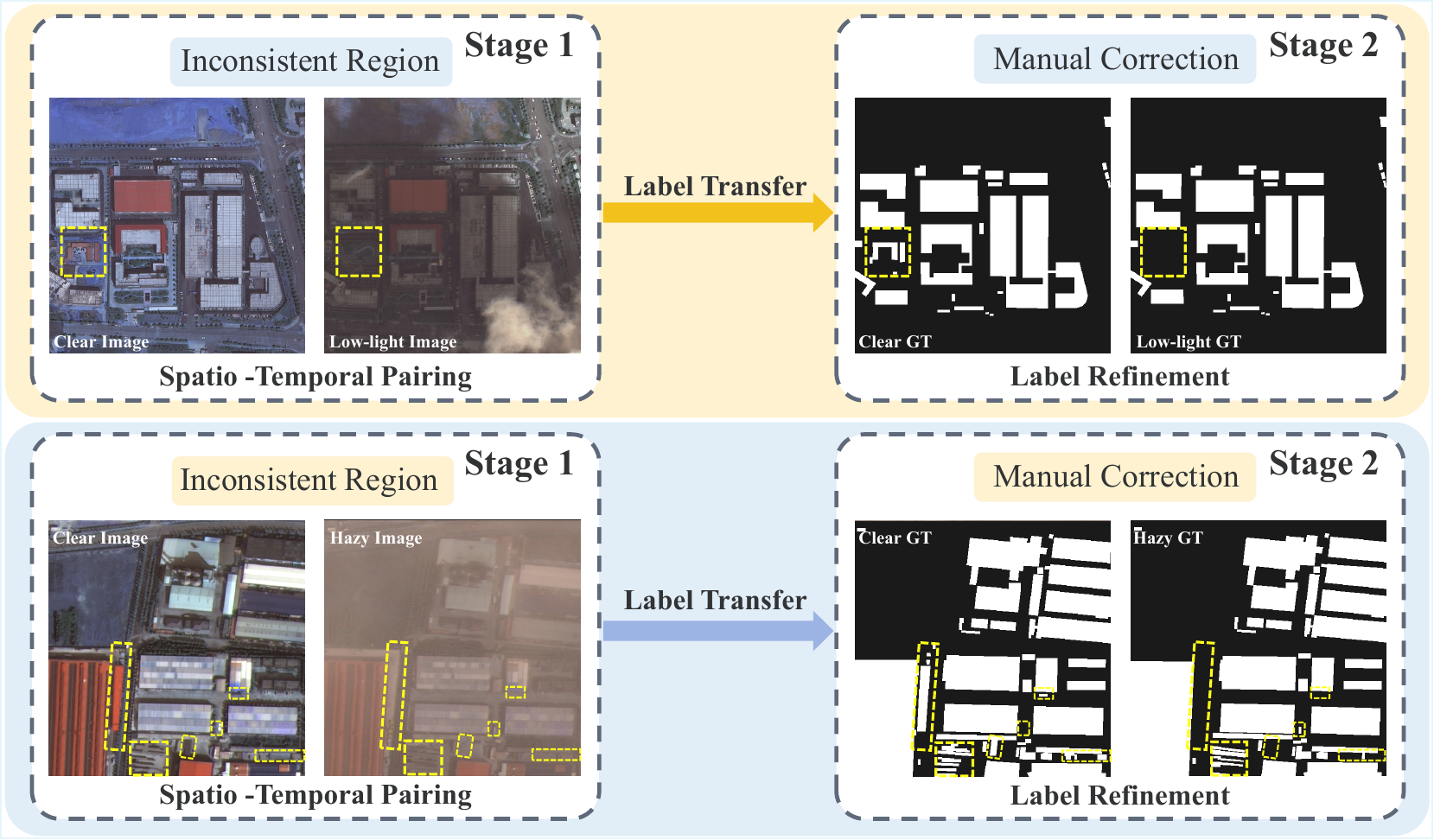}\vspace{-2mm}
	\caption{The HaLoBuilding annotation workflow. In Stage 1, clear and degraded image pairs are constructed using Same-Scene Multitemporal Pairing. Stage 2 subsequently applies label transfer coupled with rigorous manual refinement. The yellow dashed boxes indicate specific temporal or visibility disparities that underwent manual rectification to guarantee precise pixel-level alignment.}
	\label{fig1}
\end{figure}

\begin{table*}[htbp]	\centering 	\scriptsize
	\caption{Comparison between HaLoBuilding and Existing Mainstream Remote Sensing Datasets.}\vspace{-2mm}
	\label{table1}
	\renewcommand{\arraystretch}{1} 
	\setlength{\tabcolsep}{13pt} 
	\begin{tabular}{lcccccc} 
	\toprule
	Dataset & Task & Modality & Weather & Resolution(m) & Count & Image size \\ 
	\midrule
	WHU Building~\cite{whudataset} & Building Extraction & Optical & Clear & 0.3 & 8189 & $512 \times 512$ \\
	INRIA~\cite{inriadataset} & Building Extraction & Optical & Clear & 0.3 & 360 & $5000 \times 5000$ \\
	Massachusetts~\cite{Massachusetts} & Building Extraction & Optical & Clear & 1 & 151 & $1500 \times 1500$ \\
	SpaceNet 6~\cite{9150641} & Building Extraction & SAR+Optical & All-Weather & $0.5\sim2$ & 3401 & $900 \times 900$ \\
	SpaceNet 7~\cite{9578467} & Building Extraction & Optical & Clear & 4 & 2389 & $1024 \times 1024$ \\
	OpenEarthMap~\cite{10030160} & Semantic Segmentation & Optical & Clear & $0.25\sim0.5$ & 5000 & $1024 \times 1024$ \\
	LoveDA~\cite{LoveDA} & Semantic Segmentation & Optical & Clear & 0.3 & 5987 & $1024 \times 1024$ \\
	UAVid~\cite{UAVid} & Semantic Segmentation & Optical & Clear & - & 300& $\sim 4000 \times 2160$ \\
	ISPRS Vaihingen~\cite{2014ISPRS} & Semantic Segmentation & Optical & Clear & 0.09 & 33 & $2500 \times 2500$ \\
	ISPRS Potsdam~\cite{2014ISPRS} & Semantic Segmentation & Optical & Clear & 0.05 & 38 & $6000 \times 6000$ \\ 
	HRSI~\cite{Hazy_10658989} & Object Detection & Optical & Haze & $0.7\sim1.78$ & 796 & $512^2 \sim 4000^2$  \\
	RRSHID~\cite{11058953} & Image Dehazing & Optical & Haze & 1 & 3053 & $256 \times 256$ \\
	\textbf{HaLo-L (Ours)} & Building Extraction & Optical & \textbf{Low-light} & 0.8 & 2514 & $1024 \times 1024$ \\
	\textbf{HaLo-H (Ours)} & Building Extraction & Optical & \textbf{Haze} & 0.8 & 1872 & $1024 \times 1024$ \\
	\bottomrule
	\end{tabular}
\end{table*}

\begin{figure}[t]
	\centering 
	\includegraphics[width=0.48\textwidth]{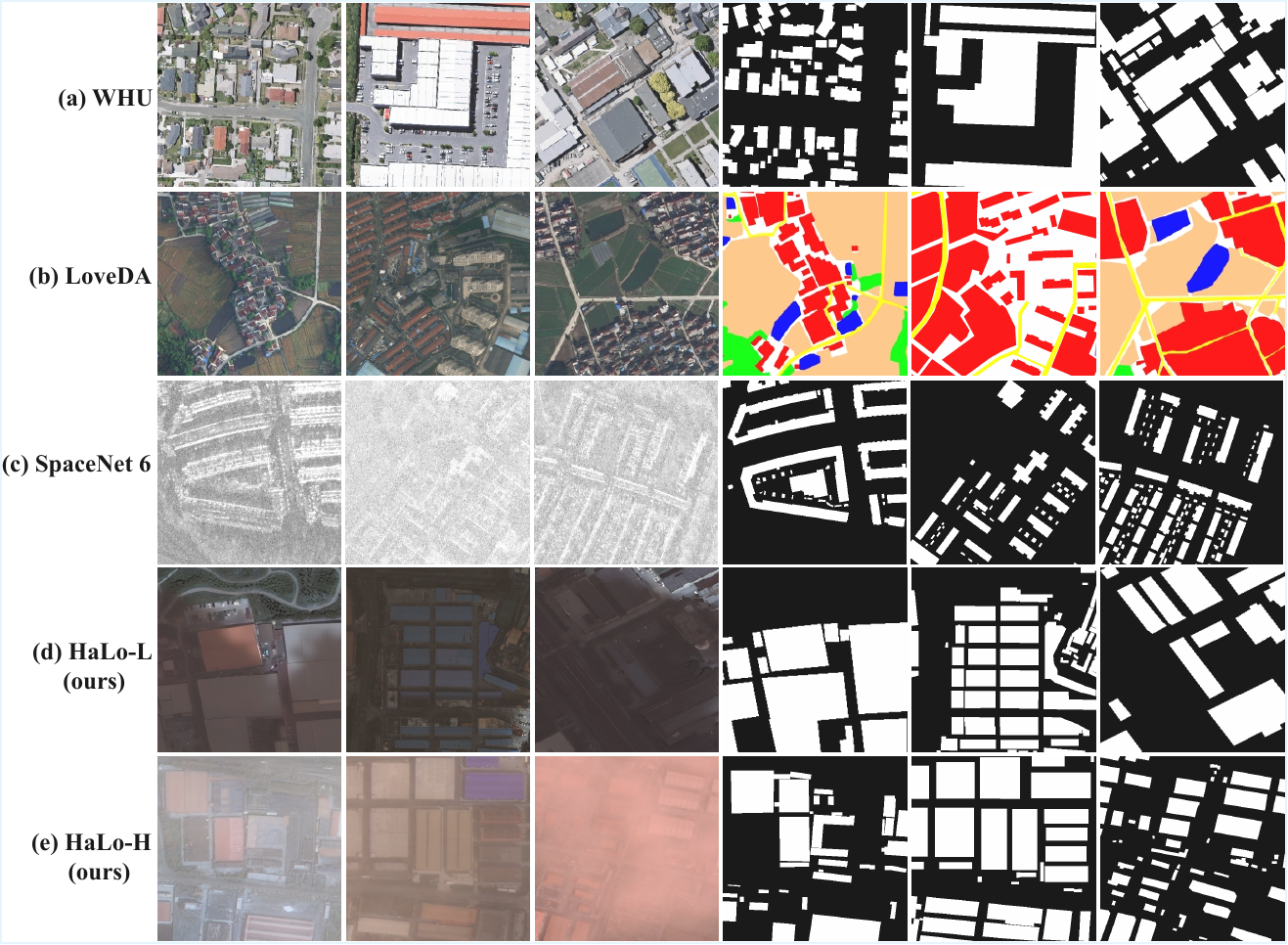}\vspace{-2mm}
	\caption{Sample images across different datasets. While (a) WHU and (b) LoveDA provide conventional clear-weather views, (c) SpaceNet 6 suffers from inherent SAR distortions. To address real-world optical degradations, (d) HaLo-L depicts severe low-light conditions, and (e) HaLo-H captures diverse haze densities.}
	\label{fig2}
\end{figure}

\section{Dataset}
Although deep learning has achieved excellent performance in remote sensing building extraction, existing mainstream benchmarks focus on clear-weather conditions. This limits model generalization in real-world scenarios such as low-light and haze, making it difficult to meet all-weather requirements for nighttime urban planning and emergency response. To fill this gap, we introduce HaLoBuilding, the first benchmark dataset specifically designed for building extraction under hazy and low-light conditions.

The raw imagery of this dataset is sourced from the GF2-PMS and GF7-DLC optical satellite system. As illustrated in Fig.~\ref{location}(a), the dataset spans 2021 to 2023 and covers diverse scenes including urban, rural, and coastal areas across provinces such as Anhui, Gansu, Ningxia, Shaanxi, Hebei, Shandong, Jiangsu, Hunan, Hubei, and Yunnan in China. 

To ensure pixel-level semantic accuracy under degraded conditions, we implemented a two-stage annotation pipeline based on the Same-Scene Multitemporal Pairing strategy, as illustrated in Fig.~\ref{fig1}. In Stage 1 (Spatio-Temporal Pairing), based on geographic coordinate metadata, we perform precise alignment and cropping to construct image pairs. Each pair consists of an adverse-weather image and a corresponding clear reference image captured within a 1-to-3-month interval. Subsequently, all aligned image pairs are cropped and standardized into $1024 \times 1024$ RGB patches. In Stage 2 (Label Refinement), we first delineate high-quality labels on the clear reference images and transfer them to the adverse-weather images. However, as indicated by the yellow dashed boxes in the figure, inconsistent regions inevitably arise due to temporal changes or visibility degradation. To address this, we conducted rigorous Manual Correction. This process ensures that the final ground truth maintains high-fidelity, pixel-level alignment with the actually visible buildings.

HaLoBuilding comprises two subsets: HaLo-L and HaLo-H, totaling $4386$ images. For fair and reproducible evaluation, we split the data in a 7:1:2 ratio for training, validation, and testing. The HaLo-L (Low-light) subset contains $2514$ images covering a range of intensities from dim illumination to near-complete darkness, challenging models to handle low signal-to-noise ratio, texture loss, and halo interference. The HaLo-H (Haze) subset includes $1872$ real hazy images spanning thin to dense fog, testing model context reasoning under low contrast, color distortion, and non-uniform scattering. The core challenge of HaLoBuilding is fundamentally rooted in a profound statistical domain shift. We conducted Kernel Density Estimation (KDE) to analyze the pixel intensity distributions of these subsets. As quantitatively revealed in Fig.~\ref{location}(b) and (c), compared to clear-weather imagery, the HaLo-H subset exhibits a distinct rightward shift accompanied by a flatter distribution (peak attenuation), quantitatively revealing the elevation of background luminance and the loss of feature contrast caused by atmospheric scattering. Conversely, the HaLo-L subset shows a severe leftward skew with intensity compression toward the lower end, confirming the drastic information loss under extreme low-light conditions. This fundamental divergence in data distribution implies that models supervised solely on ideal clear-weather benchmarks are bound to encounter catastrophic generalization failure when confronted with such extreme shifts. 

Table~\ref{table1} compares the HaLoBuilding dataset with existing mainstream remote sensing datasets. As shown in the table, classic datasets like WHU Building, INRIA, and Massachusetts primarily focus on building extraction, while LoveDA, UAVid, and ISPRS datasets emphasize multi-class semantic segmentation tasks. However, these datasets are predominantly collected under ideal clear-weather conditions. Although recent global-scale benchmarks such as SpaceNet 7~\cite{9578467} and OpenEarthMap~\cite{10030160} have significantly expanded geographic coverage, they still largely overlook the complex atmospheric environments encountered in practical applications. Fig.~\ref{fig2} illustrates the imaging differences across various datasets. To address these optical limitations, recent research has introduced real-world degraded datasets such as HRSI~\cite{Hazy_10658989} for object detection and RRSHID~\cite{11058953} for image dehazing. While these efforts effectively mitigate the domain gap between synthetic models and real physical environments, they are not specifically tailored for building extraction. Furthermore, while multi-modal datasets like SpaceNet 6~\cite{9150641} introduce SAR data as an all-weather alternative, SAR imagery inherently suffers from radiometric noise and severe geometric distortions. As demonstrated in Fig.~\ref{fig2}(c), the side-looking geometry causes a significant displacement between the building's top and its actual geographic footprint due to layover effects, making precise boundary localization extremely challenging. Consequently, a dedicated large-scale optical benchmark for building extraction in both hazy and low-light scenarios remains absent. HaLoBuilding is designed to bridge this critical gap. By focusing specifically on building extraction in hazy and low-light conditions, it provides a rigorous and high-fidelity benchmark that directly addresses the unique challenges of hazy and low-light conditions.

\begin{figure*}[t]
	\centering 
	\includegraphics[width=0.9\textwidth]{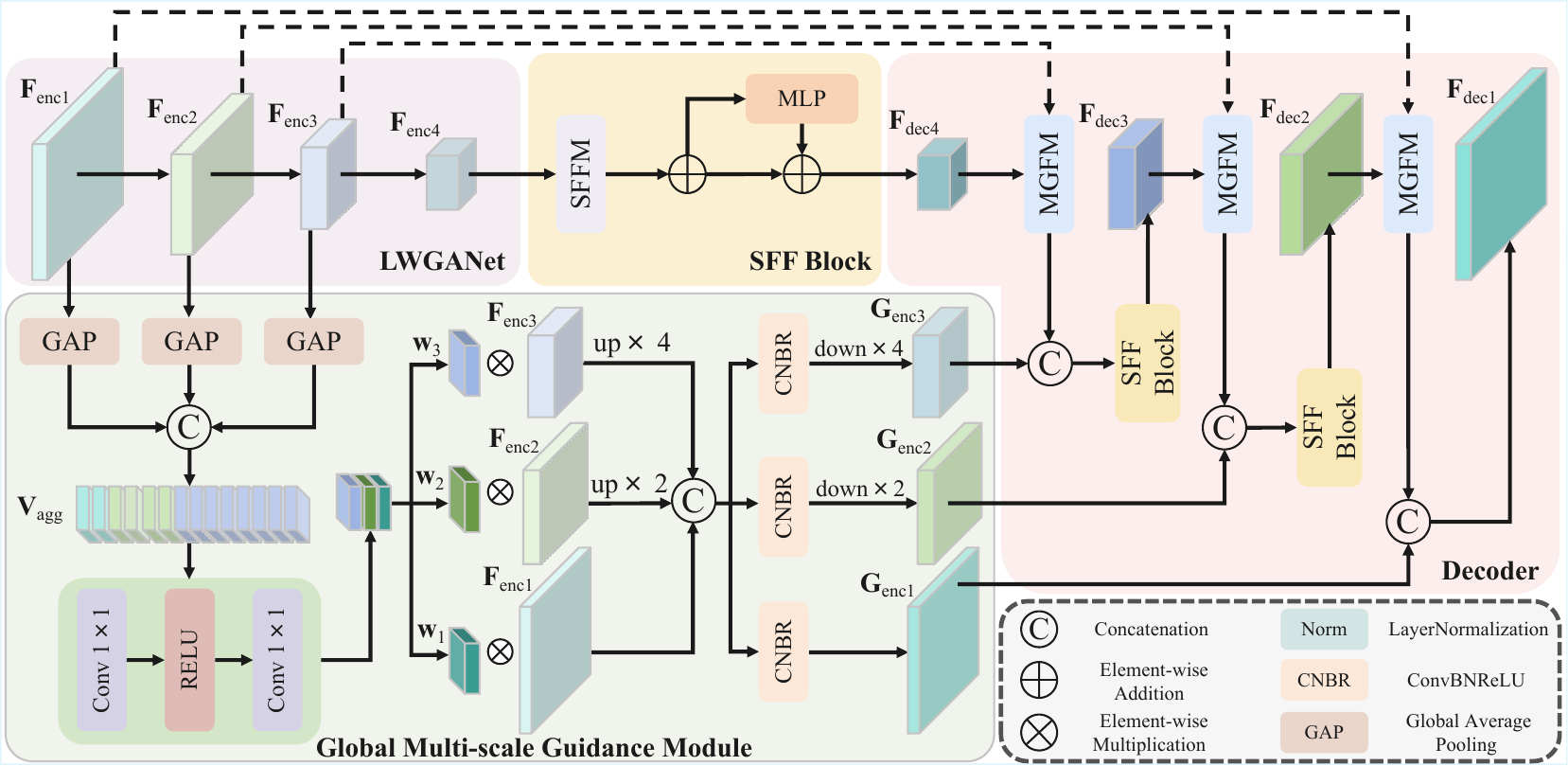}\vspace{-2mm}
	\caption{Overall architecture of the proposed HaLoBuild-Net and the structure of the Global Multi-scale Guidance Module (GMGM). Following hierarchical feature extraction by the lightweight LWGANet-L2 encoder, the decoder progressively reconstructs building contours through three core components: the GMGM aggregates multi-scale shallow features to provide global semantic constraints, effectively anchoring building topologies even under severe degradation; the MGFM performs complex bidirectional semantic-spatial calibration; and the SFF Block combines the SFFM and MLP paths for collaborative dual-domain optimization.}
	\label{fig3}
\end{figure*}

\begin{figure} 
	\centering
	\includegraphics[width=\linewidth]{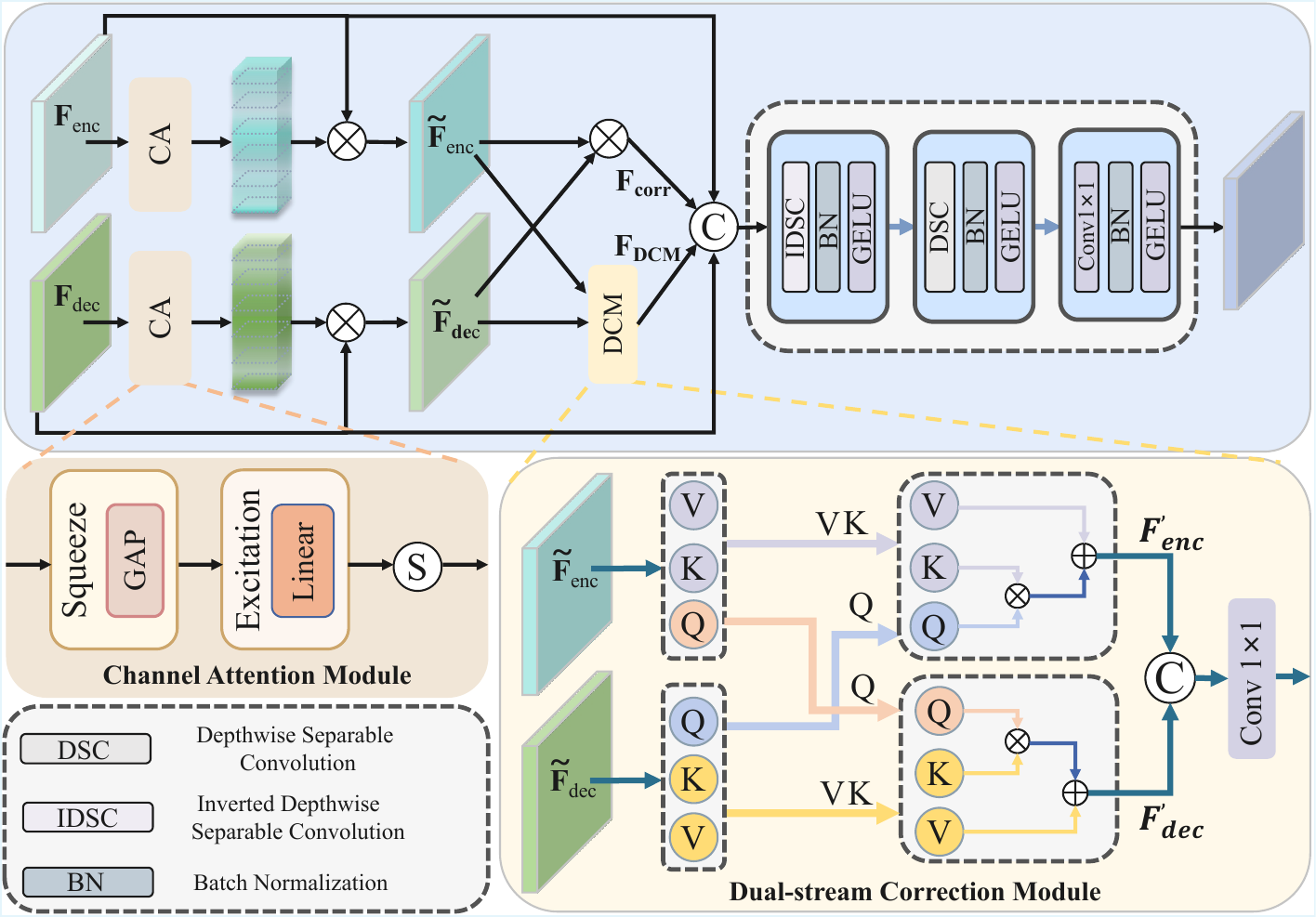}
	\caption{Structure of the MGFM. MGFM is designed for bidirectional semantic-spatial calibration. It uses deep semantics as a spatial filter to suppress weather-induced noise in shallow layers, while leveraging shallow spatial anchors to refine blurred deep-layer boundaries.}
	\label{fig4}
\end{figure}

\section{Methodology}

\subsection{Overview}
\label{4.1}
As illustrated in Fig.~\ref{fig3}, HaLoBuild-Net adopts a robust end-to-end encoder-decoder architecture to directly process hazy and low-light imagery without explicit enhancement. Following hierarchical feature extraction by the lightweight LWGANet backbone~\cite{lu2025lwganet}, the decoder progressively reconstructs building contours through a cascaded refinement scheme. This process synergistically integrates MGFM for bidirectional semantic-spatial calibration, GMGM for global topological anchoring, and SFFM for frequency-aware feature modulation. By leveraging the empirical stability of low-frequency spectral anchors and the mutual guidance between these hierarchical features, HaLoBuild-Net collectively ensures high-fidelity extraction even under complex degradations.

\subsection{Encoder}
\label{4.2}
We select the lightweight backbone LWGANet-L2~\cite{lu2025lwganet}, specifically designed for remote sensing tasks, as the feature extractor. Its core Lightweight Group Attention mechanism enables simultaneous capture of information from small, medium, and large-scale targets with low computational overhead, effectively recognizing buildings of varying sizes under hazy and low-light conditions. The encoder takes the input image and outputs four feature maps at different resolutions, corresponding to 1/4, 1/8, 1/16, and 1/32 of the original image resolution.

\subsection{MGFM}
\label{4.3}
Standard skip connections often suffer from feature contamination in adverse weather. Specifically, haze causes deep-layer boundary diffusion, while low-light corrupts shallow features with sensor noise. To resolve this, we propose the Mutual-Guided Fusion Module (MGFM) for bidirectional semantic-spatial calibration. MGFM uses deep semantics as a spatial filter to suppress weather-induced noise in shallow layers, while leveraging shallow spatial anchors to refine blurred deep-layer boundaries, effectively decoupling building structures from environmental interference. The structure of MGFM is illustrated in Fig.~\ref{fig4}.

First, the input encoder feature $F_{\mathrm{enc}}$ and decoder feature $F_{\mathrm{dec}}$ undergo channel attention preprocessing to adaptively recalibrate feature responses:
\begin{align}
    \tilde{F}_{\mathrm{enc}} &= F_{\mathrm{enc}} \odot \sigma(\mathrm{Linear}(\mathrm{GAP}(F_{\mathrm{enc}}))), \label{eq:enc_att} \\
    \tilde{F}_{\mathrm{dec}} &= F_{\mathrm{dec}} \odot \sigma(\mathrm{Linear}(\mathrm{GAP}(F_{\mathrm{dec}}))), \label{eq:dec_att}
\end{align}
\noindent where $\sigma(\cdot)$ is the Sigmoid activation, $\mathrm{Linear}(\cdot)$ is a fully connected layer, and $\mathrm{GAP}(\cdot)$ represents global average pooling. $\odot$ denotes element-wise multiplication.

After enhancement, MGFM processes features via two parallel paths. The first is the Dual-stream Correction Module (DCM), which employs cross-attention for bidirectional guidance. Specifically, queries $Q_{\mathrm{dec}}$ from deep semantics are used to filter meteorological noise in shallow layers, while queries $Q_{\mathrm{enc}}$ from shallow details retrieve spatial anchors to sharpen the blurred boundaries of deep features. The corrected features are~computed as:
\begin{align}
    F'_{\mathrm{enc}} &= V_{\mathrm{enc}} + (Q_{\mathrm{dec}} \odot K_{\mathrm{enc}}), \label{eq:dcm_enc} \\
    F'_{\mathrm{dec}} &= V_{\mathrm{dec}} + (Q_{\mathrm{enc}} \odot K_{\mathrm{dec}}), \label{eq:dcm_dec}
\end{align}
\noindent where $Q$, $K$, and $V$ denote query, key, and value vectors, respectively. The outputs are then fused to yield $F_{\mathrm{DCM}}$:
\begin{equation}
    F_{\mathrm{DCM}} = \mathrm{Conv}_{1 \times 1}(\mathrm{Concat}_c(F'_{\mathrm{enc}}, F'_{\mathrm{dec}})), \label{eq:fdcm}
\end{equation}
\noindent where $\mathrm{Concat}_c(\cdot)$ denotes channel-wise concatenation.

Simultaneously, the second path computes an explicit correlation feature $F_{\mathrm{corr}}$ to preserve original building information in high-confidence regions. Finally, a global residual fusion strategy integrates all cues:
\begin{align}
    F_{\mathrm{corr}} &= \tilde{F}_{\mathrm{enc}} \odot \tilde{F}_{\mathrm{dec}}, \label{eq:fcorr} \\
    F_{\mathrm{cat}} &= \mathrm{Concat}_c(F_{\mathrm{enc}}, F_{\mathrm{dec}}, F_{\mathrm{corr}}, F_{\mathrm{DCM}}). \label{eq:fcat}
\end{align}

The concatenated feature is processed by a fusion block (including IDSC and DSC) for dimensionality reduction. By utilizing deep semantics to suppress shallow noise and leveraging shallow details to compensate for weather-induced boundary diffusion, MGFM enables high-fidelity building reconstruction under complex degradations.

\subsection{GMGM}
\label{4.4}
In degraded images under hazy and low-light conditions, local features are often highly uncertain, and relying solely on local receptive fields can easily lead to misjudgments. To this end, we design the GMGM, which jointly analyzes multi-scale shallow features to generate high-level semantic planning maps, providing global context constraints for the decoder. Its structure is shown in Fig.~\ref{fig3}.

GMGM takes the three-scale shallow encoder features $\{ F_{\mathrm{enc1}}, F_{\mathrm{enc2}}, F_{\mathrm{enc3}} \}$ as input. First, global average pooling is applied to each scale feature, followed by channel-wise concatenation to form the global context descriptor $V_{\mathrm{agg}}$:
\begin{equation}
    V_{\mathrm{agg}} = \mathrm{Concat}_c \left( \{ \mathrm{GAP}(F_{\mathrm{enc}i}) \}_{i=1}^{3} \right),
    \label{eq:vagg}
\end{equation}
\noindent where $i \in \{1, 2, 3\}$ denotes the scale index.

Subsequently, $V_{\mathrm{agg}}$ is fed into a lightweight cross-scale attention network to adaptively generate scale weights $\{ w_1, w_2, w_3 \}$. The network consists of two $1 \times 1$ convolutions and a ReLU activation $\delta$. This process can be formulated as:
\begin{equation}
    \{w_1, w_2, w_3\} = \mathrm{Split}(\mathrm{Conv}_{1 \times 1}(\delta(\mathrm{Conv}_{1 \times 1}(V_{\mathrm{agg}})))), \label{eq:weights}
\end{equation}
\noindent where $\mathrm{Split}(\cdot)$ denotes channel splitting and $\delta(\cdot)$ is ReLU.

Subsequently, the weighted features are aligned according to their scale discrepancies. Specifically, the mid- and low-resolution features undergo two-fold and four-fold upsampling, respectively, before being concatenated with the highest-resolution feature to yield feature $F_{\mathrm{global}}$:
\begin{small} 
	\begin{equation}
		F_{\mathrm{global}} = \mathrm{Concat}_c \big( w_1 F_{\mathrm{enc1}}, U(w_2 F_{\mathrm{enc2}}), U(w_3 F_{\mathrm{enc3}}) \big),
		\label{eq:ffused}
	\end{equation}
	\end{small}
\noindent where $U(\cdot)$ represents bilinear interpolation upsampling.

Finally, the feature $F_{\mathrm{global}}$ is projected through three parallel ConvBNReLU (CNBR) branches and then downsampled to adjust back to the original scales, yielding guidance features $\{ G_{\mathrm{enc1}}, G_{\mathrm{enc2}}, G_{\mathrm{enc3}} \}$:
\begin{align}
    G_{\mathrm{enc1}} &= \mathrm{ConvBNReLU}_1(F_{\mathrm{global}}), \label{eq:genc1} \\
    G_{\mathrm{enci}} &= D_i(\mathrm{ConvBNReLU}_i(F_{\mathrm{global}})), \quad i \in \{2, 3\}, \label{eq:genci}
\end{align}
\noindent where $\mathrm{ConvBNReLU}_i(\cdot)$ is the $i$-th Conv-BN-ReLU branch and $D_i(\cdot)$ represents downsampling to the $i$-th original scale. Enriched with global-aware semantic context, these guidance features are integrated into the features of each decoder stage via concatenation. This process provides explicit spatial priors to anchor building topologies, ensuring that the model can accurately reconstruct structural boundaries even when deep features are obscured by degradation-induced blurring.
\begin{figure} 
	\centering
	\includegraphics[width=\linewidth]{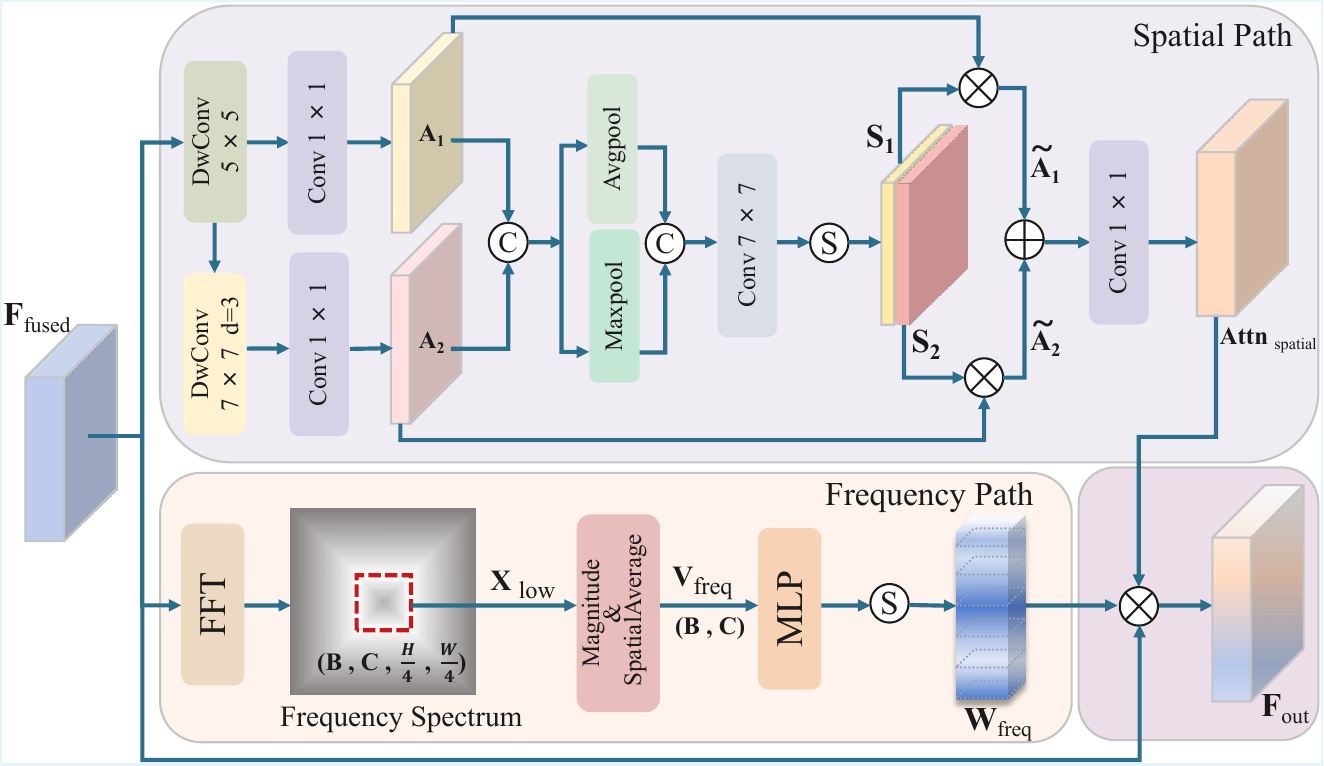}\vspace{-2mm}
	\caption{Structure of the SFFM. SFFM achieves dual-domain collaborative optimization. It couples a large receptive field spatial attention branch (to repair geometric discontinuities) with a frequency-aware channel reweighting mechanism guided by stable low-frequency anchors (to suppress unreliable high-frequency noise).}
	\label{fig5}
\end{figure}

\subsection{SFFM}
\label{4.5}
At each decoder stage, the fused features require further structural enhancement. To this end, we design the SFF Block as the fundamental processing unit. As illustrated in the yellow box of Fig.~\ref{fig3}, each SFF Block follows a residual learning structure: it first employs the SFFM to perform collaborative optimization in both spatial and frequency domains, followed by a Multi-Layer Perceptron (MLP) for channel-wise feature projection and non-linear mapping. Normalization (Norm) and residual additions are integrated to ensure training stability and facilitate gradient flow. The detailed internal structure of the SFFM is further elaborated in Fig.~\ref{fig5}.

In the spatial domain, we construct a large receptive field attention branch to repair geometric discontinuities caused by hazy and low-light conditions. The input feature $F_{\mathrm{fused}} \in \mathbb{R}^{B \times C \times H \times W}$ first undergoes a $5 \times 5$ Depthwise Separable Convolution and a $1 \times 1$
 convolution to generate preliminary feature $A_1$. Subsequently, $A_1$ is processed by a $7 \times 7$ dilated Depthwise Separable Convolution with a dilation rate of $3$ to expand the receptive field, followed by a $1 \times 1$ convolution to yield feature $A_2$. The two are concatenated along the channel dimension to form feature $A$:
\begin{align}
    A_1 &= \mathrm{Conv}_{1 \times 1}(\mathrm{DWConv}_{5 \times 5}(F_{\mathrm{fused}})), \label{eq:a1} \\
    A_2 &= \mathrm{Conv}_{1 \times 1}(\mathrm{DWConv}_{7 \times 7, d=3}(A_1)), \label{eq:a2} \\
    A &= \mathrm{Concat}_c(A_1, A_2), \label{eq:a}
\end{align}
\noindent where $\mathrm{DWConv}_{k \times k}(\cdot)$ denotes Depthwise Separable Convolution with kernel size $k \times k$, and $d$ is the dilation rate. Specifically, $A_1$ captures fine-grained local details, while the dilated branch $A_2$ incorporates multi-scale context, enabling the perception of diverse building footprints under degraded visibility.

Subsequently, average and max pooling are applied simultaneously along the channel dimension to extract spatial statistical descriptors. This descriptor is processed through a $7 \times 7$ convolution and Sigmoid activation to generate a dual-channel spatial attention map $M_{\mathrm{spatial}} \in \mathbb{R}^{B \times 2 \times H \times W}$:
\begin{equation}
    \resizebox{0.88\columnwidth}{!}{
        $M_{\mathrm{spatial}} = \sigma(\mathrm{Conv}_{7 \times 7}(\mathrm{Concat}_c(\mathrm{Avg}(A), \mathrm{Max}(A))))$},
    \label{eq:mspatial}
\end{equation}
\noindent where $\sigma(\cdot)$ is the Sigmoid activation, and $\mathrm{Avg}(\cdot)$ and $\mathrm{Max}(\cdot)$ denote average and max pooling operations, respectively.

The attention map is split into two single-channel maps $S_1$ and $S_2$, which recalibrate $A_1$ and $A_2$ respectively. The recalibrated results are summed and fused via $1 \times 1$ convolution to obtain the spatial attention weight $\mathrm{Attn}_{\mathrm{spatial}}$:
\begin{equation}
    \mathrm{Attn}_{\mathrm{spatial}} = \mathrm{Conv}_{1 \times 1}(A_1 \odot S_1 + A_2 \odot S_2). \label{eq:attn_spatial}
\end{equation}

In the frequency domain, we design a frequency-aware channel reweighting branch guided by low-frequency statistics. Although haze and low-light degradations arise from different physical processes, both lead to reduced reliability of high-frequency components—either due to attenuation (haze) or noise contamination (low-light). Therefore, instead of explicitly suppressing high-frequency components, we leverage the stability of low-frequency structures to guide adaptive feature modulation. Specifically, a 2D Fast Fourier Transform (FFT) is first applied to the input feature $F_{\mathrm{fused}}$, followed by a spectrum centering operation to rearrange the frequency components. A central low-frequency sub-band is then extracted:
\begin{equation}
    X_{\mathrm{low}} = \mathrm{Crop}_{\mathrm{LF}}(\mathrm{FFTShift}(\mathrm{FFT}(F_{\mathrm{fused}}))), \label{eq:xlow}
    \end{equation}
\noindent where $\mathrm{FFTShift}(\cdot)$ denotes spectrum centering and $\mathrm{Crop}_{\mathrm{LF}}(\cdot)$ extracts the centered low-frequency region.

Next, the spatial average of the magnitude spectrum of $X_{\mathrm{low}}$ is computed to obtain a channel-wise frequency descriptor $V_{\mathrm{freq}}$:
\begin{equation}
    V_{\mathrm{freq}} = \mathrm{Mean}_{\mathrm{sp}}(|X_{\mathrm{low}}|), \label{eq:vfreq}
\end{equation}
\noindent where $|\cdot|$ denotes magnitude and $\mathrm{Mean}_{\mathrm{sp}}(\cdot)$ denotes averaging over the spatial dimension.

Since low-frequency components are more robust to noise and preserve global structural information, the descriptor $V_{\mathrm{freq}}$ serves as a stable prior for feature modulation. It is further passed through a lightweight multi-layer perceptron (MLP) followed by a Sigmoid activation to generate channel-wise frequency attention weights $W_{\mathrm{freq}} \in \mathbb{R}^{B \times C \times 1 \times 1}$:
\begin{equation}
    W_{\mathrm{freq}} = \sigma(\mathrm{MLP}(V_{\mathrm{freq}})). \label{eq:wfreq}
\end{equation}

This design enables the model to adaptively emphasize reliable structural components while suppressing unreliable responses under different degradation conditions.

Finally, under the collaborative modulation of spatial attention and frequency-aware channel weights, the input feature is~adaptively refined:
\begin{equation}
    F_{\mathrm{out}} = F_{\mathrm{fused}} \odot \mathrm{Attn}_{\mathrm{spatial}} \odot W_{\mathrm{freq}}. \label{eq:frefined}
\end{equation}

This dual-domain mechanism combines spatial edge sharpening with frequency-domain structural anchoring. By integrating these complementary cues, SFFM achieves both sharp boundaries and topological integrity, effectively resolving the trade-off between noise suppression and detail preservation.


\section{Experiments}

\subsection{Experimental Datasets}
To evaluate the effectiveness and generalization of HaLoBuild-Net across diverse scenarios, we conduct experiments on the following four datasets:

\begin{itemize}
    \item \textbf{HaLoBuilding (Ours):} Our proposed benchmark specifically designed for building extraction under adverse scenarios. It contains 4,386 high-resolution image pairs from multiple  provinces in China, divided into \textbf{HaLo-H} (Haze) and \textbf{HaLo-L} (Low-light). We adopt a 7:1:2 ratio for training, validation, and testing.
    
    \item \textbf{WHU Building Dataset}~\cite{whudataset}: A widely used high-precision aerial dataset with 0.3~m resolution. We utilize its aerial subset containing 8,189 images of $512 \times 512$ pixels to evaluate the model's performance in fine-grained extraction under clear-weather conditions.
    
    \item \textbf{INRIA Aerial Image Labeling Dataset}~\cite{inriadataset}: This dataset covers diverse urban landscapes across five cities in Europe and the United States ($5000 \times 5000$ tiles at 0.3~m resolution). It is primarily employed to assess the cross-region generalization capability of the models.
    
    \item \textbf{LoveDA Dataset}~\cite{LoveDA}: Consisting of 5,987 images ($1024 \times 1024$ pixels) from three Chinese cities, this dataset features complex backgrounds in both urban and rural scenes, making it ideal for verifying building separation in heterogeneous environments.
\end{itemize}

\subsection{Evaluation Metrics}
To quantitatively evaluate the performance of our proposed model and all comparison methods, we adopt four widely used metrics in semantic segmentation tasks: Intersection over Union (IoU), F1-Score (F1), Precision (Pre), and Recall.

\subsection{Implementation Details}
All experiments are conducted on an NVIDIA RTX 3090 Ti ($24$~GB) GPU, with a fixed random seed of $42$ for reproducibility. Input images are originally $1024 \times 1024$, subjected to random scaling and flipping, then cropped to $512 \times 512$ for training. Models are trained for 100 epochs using the AdamW optimizer~\cite{loshchilov2018decoupled} coupled with the Lookahead strategy, utilizing a batch size of 16, an initial learning rate of $4 \times 10^{-4}$, and a cosine annealing decay scheduler with a weight decay of 0.01. The loss function integrates Soft Cross-Entropy and Dice loss to simultaneously jointly optimize pixel-level accuracy and regional overlap.

\subsection{Comparison with State-of-the-Art Methods}
\begin{table*}[t]
	\centering  
	\scriptsize
	\caption{Comparison Results on HaLo-L and HaLo-H Datasets. The Best, Second-best, and Third-best Results are Highlighted in \textcolor{red}{red}, \textcolor{blue}{blue}, and \textcolor{green}{green}. Additionally, IoU Denotes the Intersection over Union for the Building Class. }\vspace{-2mm}
	\label{tab:final_comparison}
	\renewcommand{\arraystretch}{1.1} 
	\setlength{\tabcolsep}{10pt} 
	\begin{tabular}{l c cc cccc cccc}
	\toprule
	\multirow{2}{*}{Method} & \multirow{2}{*}{Publication} 
	& \multirow{2}{*}{\shortstack[c]{Params.  \\ (M) $\downarrow$}} 
	& \multirow{2}{*}{\shortstack[c]{FLOPs  \\ (G)$\downarrow$}}  
	& \multicolumn{4}{c}{HaLo-L (\%) $\uparrow$} 
	& \multicolumn{4}{c}{HaLo-H (\%) $\uparrow$} \\
	\cmidrule(r){5-8} \cmidrule(l){9-12} 
	& & & & IoU & F1 & Pre & Recall
	& IoU & F1 & Pre & Recall \\
	\midrule
	UNetFormer~\cite{Wang_2022}     & ISPRS 2022 & 11.7 & 46.9 & 66.40 & 79.81 & \textcolor{blue}{84.30} & 75.78 & 67.92 & 80.90 & 86.74 & 75.79 \\
	Buildformer~\cite{wang2022building}    & TGRS 2022  & 40.5 & 116.6 & 62.90 & 77.23 & 82.47 & 72.62 & 62.76 & 77.12 & 86.83 & 69.36 \\
	SACANet~\cite{ma2023sacanet}        & ICME 2023  & 2.7 & 44.4 & 64.17 & 78.17 & \textbf{\textcolor{red}{86.73}} & 71.15 & 54.55 & 70.59 & \textcolor{green}{89.05} & 58.47 \\
	EasyNet~\cite{10376153}        & TGRS 2024  & 3.9 & 17.3 & 61.56 & 72.79 & 77.39 & 75.13 & 57.37 & 68.96 & 74.47 & 70.81 \\
	UANet~\cite{10418227}          & TGRS 2024  & 15.6 & 27.8 & 56.72 & 72.38 & 80.30 & 65.89 & 56.47 & 72.18 & \textcolor{blue}{90.56} & 60.00 \\
	DecoupleNet-D2~\cite{10685518}    & TGRS 2024  & 6.9 & 32.1 & \textcolor{green}{66.56} & \textcolor{green}{79.92} & 82.25 & \textcolor{blue}{77.72} & \textcolor{green}{68.11} & \textcolor{green}{81.03} & 85.02 & \textcolor{blue}{77.40} \\
	RSBuilding~\cite{10623867}     & TGRS 2024  & 308 & - & 66.09 & 79.58 & 82.08 & 77.23 & 67.70 & 80.74 & 85.87 & \textcolor{green}{76.19} \\
	UMformer~\cite{10969832}       & TGRS 2025  & 12.4 & 47.7 & 65.25 & 78.98 & 81.11 & 76.95 & 65.79 & 79.36 & 88.39 & 72.01 \\
	LOGCAN++~\cite{ma2025logcan}     & TGRS 2025  & 31.1 & 205.1 & 58.50 & 73.82 & 78.40 & 69.75 & 65.47 & 79.14 & \textbf{\textcolor{red}{90.80}} & 70.13 \\
	AFENet~\cite{10955240}         & TGRS 2025  & 20.2 & 102.4 & 64.38 & 78.33 & 82.69 & 74.40 & 61.47 & 76.14 & 88.96 & 66.55 \\
	LWGANet-L2~\cite{lu2025lwganet}    & AAAI 2026  & 12.6 & 48.4 & \textcolor{blue}{66.81} & \textcolor{blue}{80.11} & 82.68 & \textcolor{green}{77.68} & \textcolor{blue}{68.84} & \textcolor{blue}{81.54} & 87.93 & 76.02 \\
	\midrule
	HaLoBuild-Net (ours) & - & 23.9 & 90.3 & \textbf{\textcolor{red}{68.90}} & \textbf{\textcolor{red}{81.59}} & \textcolor{green}{84.16} & \textbf{\textcolor{red}{79.10}}
	& \textbf{\textcolor{red}{70.88}} & \textbf{\textcolor{red}{82.96}} & 86.92 & \textbf{\textcolor{red}{79.35}} \\
	\bottomrule
	\end{tabular}
	\label{table2}
\end{table*}

\begin{figure*}[t]
	\centering 
	\includegraphics[width=0.8\textwidth]{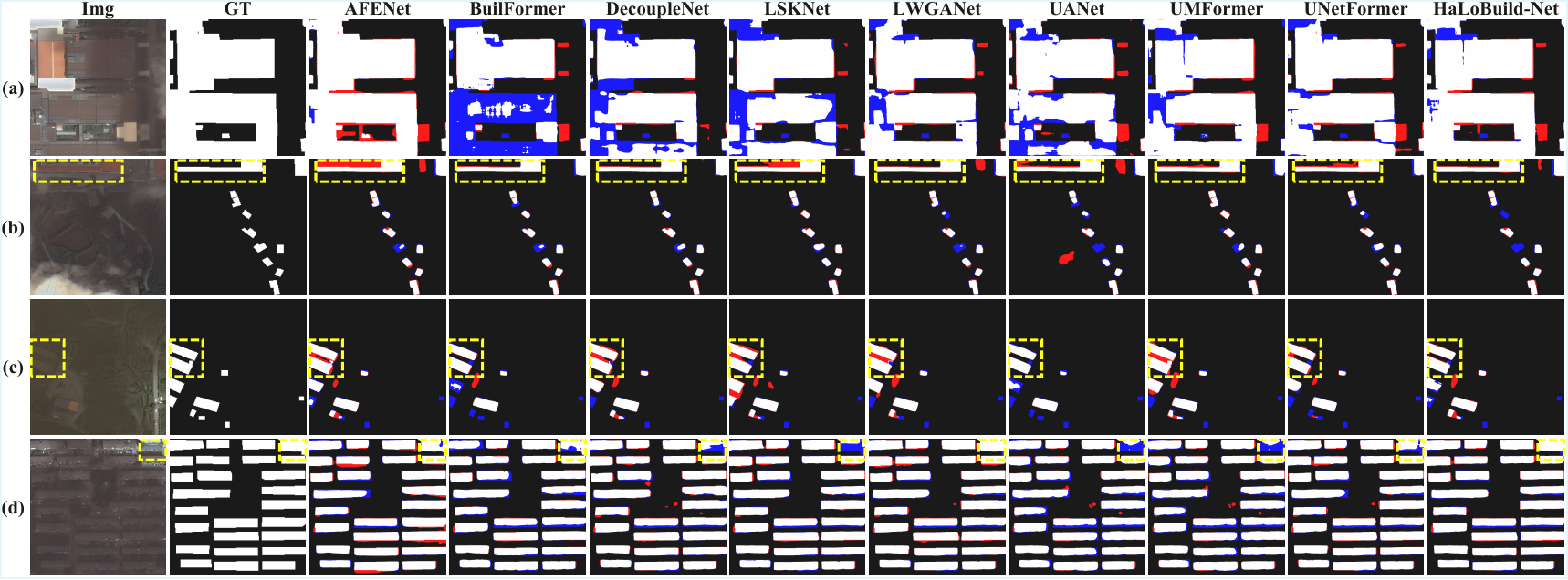}\vspace{-2mm}
	\caption{Visual comparison of building extraction results on HaLo-L dataset. It demonstrates the model's remarkable robustness under extreme low-light conditions. It successfully extracts complete large-scale structures without internal voids, precisely separates buildings from texture-similar backgrounds (like cement roads), and accurately captures small targets despite weak visual cues. Error maps: White (TP), Red (FP), and Blue (FN).}
	\label{fig6}
\end{figure*}

\subsubsection{Performance on HaLoBuilding Dataset}\label{sec:AdverseWeatherResults}\mbox{}

Quantitative comparison results on the HaLoBuilding dataset are presented in Table~\ref{table2}. On the HaLo-L subset, compared to the strongest competitor LWGANet-L2, our HaLoBuild-Net achieves superior performance in IoU, F1, and Recall, with significant improvements of $2.09\%$, $1.48\%$, and $1.42\%$, respectively. On the HaLo-H subset, HaLoBuild-Net also demonstrates state-of-the-art results. It achieves the highest IoU of $70.88\%$ and an F1 score of $82.96\%$, outperforming LWGANet-L2 by $2.04\%$ and $1.42\%$ in these two key metrics. These results validate that our proposed dual-domain collaborative architecture and multi-scale aggregation strategy effectively mitigate feature degradation caused by hazy and low-light conditions, significantly enhancing building extraction performance.

For more intuitive comparison, we use error maps (white: TP, red: FP, blue: FN). Fig.~\ref{fig6} shows visualization results on the HaLo-L test set. Specifically, (a) demonstrates HaLoBuild-Net's strong global understanding in low-contrast dark regions, enabling it to extract complete large-scale structures while effectively suppressing background noise, avoiding internal voids caused by weakened boundary information in other methods. (b) and (c) highlight its precise edge representation and strong discriminative capability in complex backgrounds, successfully separating buildings from texture-similar cement roads and grass, overcoming boundary blurring due to low light. (d) showcases its proficiency in capturing small targets, accurately locating them even under extremely weak visual cues, addressing the challenge of easy oversight during feature extraction. Overall, HaLoBuild-Net exhibits remarkable robustness under low light, effectively adapting to complex backgrounds and multi-scale variations.
\begin{figure*}[t]
    \centering 
    \includegraphics[width=0.8\textwidth]{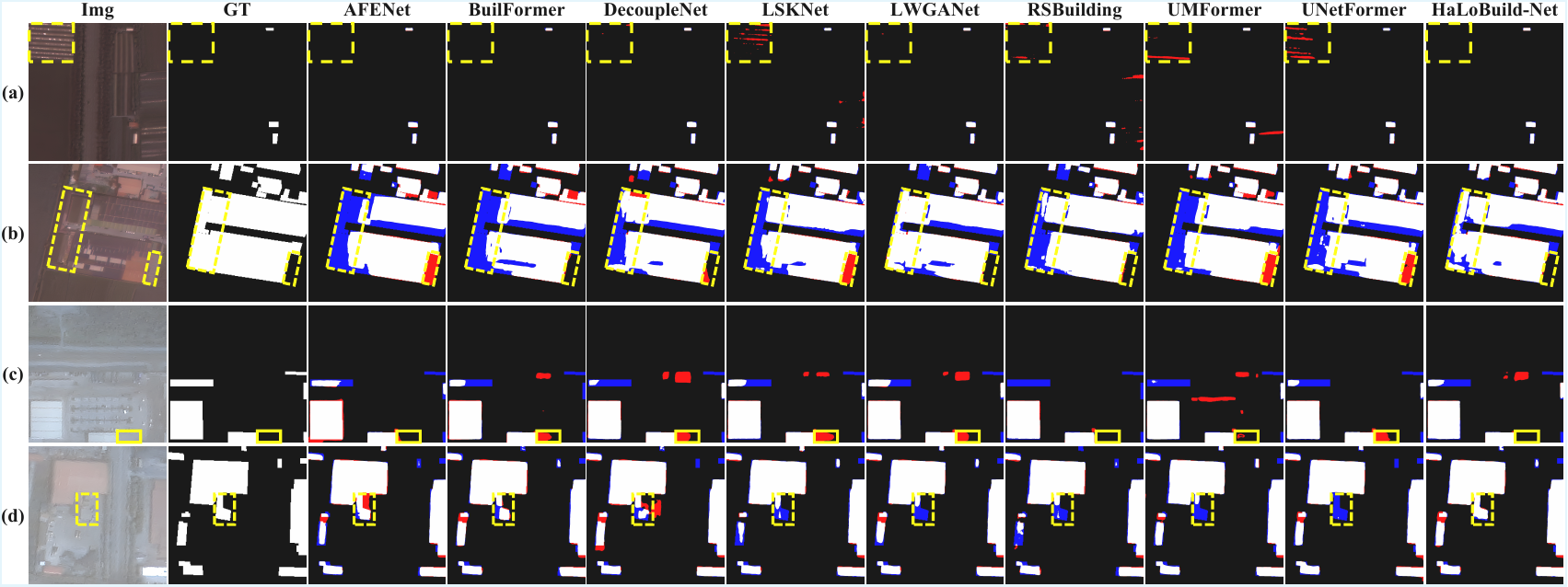}\vspace{-2mm}
    \caption{Visual comparison of building extraction results on HaLo-H dataset. It illustrates the model's strong semantic discrimination capability under dense haze. It effectively penetrates atmospheric scattering to distinguish rigid building structures from confusable features (like agricultural greenhouses), successfully separating low-contrast buildings from hazy backgrounds.}
    \label{fig7}
    \end{figure*}
Fig.~\ref{fig7} presents visualization results on the HaLo-H test set. Specifically, (a) and (c) demonstrate its strong semantic discrimination capability under dense haze interference, effectively penetrating atmospheric scattering to distinguish rigid building structures from texture-similar agricultural greenhouses and cement roads, avoiding misclassifications caused by spectral similarity in other methods. (b) highlights its superior feature representation under hazy and low-light conditions, successfully separating low-contrast dark-region buildings from hazy backgrounds, correctly identifying non-building ground. (d) illustrates the effectiveness of GMGM, which locks small target positions through strong prior injection, preventing small-scale buildings from being overwhelmed by noise in deep feature extraction and achieving precise recovery.

\begin{figure} 
	\centering
	\includegraphics[width=\linewidth]{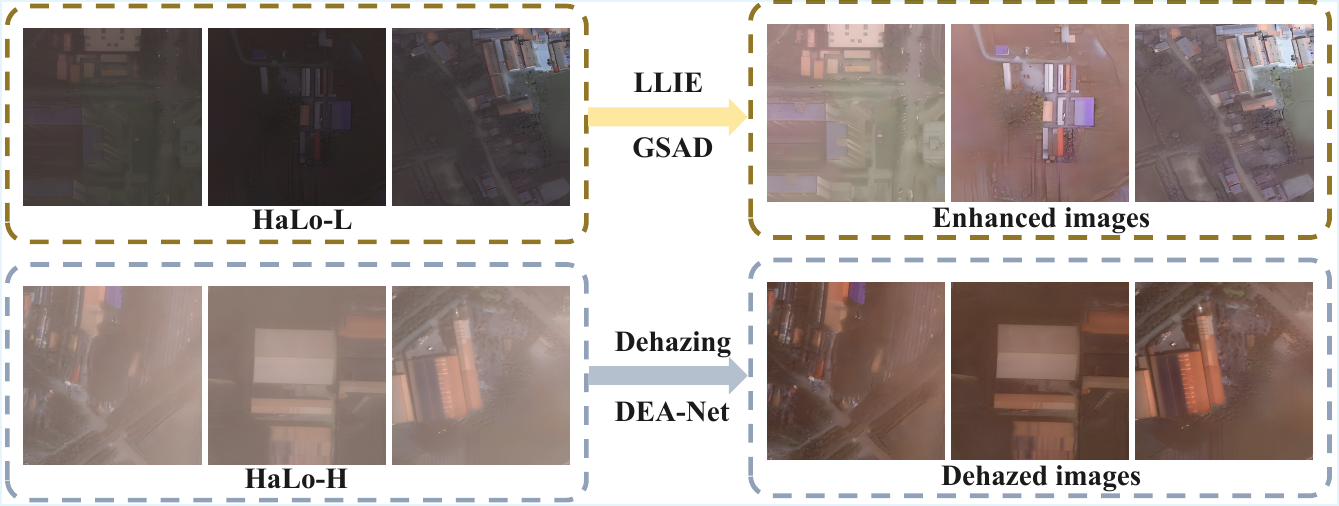}\vspace{-2mm}
	\caption{Visual demonstration of image pre-processing on the HaLoBuilding dataset. Low-Light Image Enhancement (LLIE) via GSAD on HaLo-L and image dehazing via DEA-Net on HaLo-H. These restored images serve as the pre-processed inputs for the cascaded strategy evaluation.}
	\label{enhanced}
\end{figure}

\begin{table*}[htbp]
	\centering \scriptsize \vspace{-2mm}
	\caption{Performance Comparison of Direct Versus Cascaded Building Extraction on HaLo-L and HaLo-H Datasets. Asterisk ($\ast$) Denotes the Cascaded Paradigm, where Images Undergo Low-light Enhancement and Dehazing Pre-processing before Being Fed Into the Network. Results for the Cascaded Strategy Represent the Average of Three Independent Rxperimental Runs.}
	\renewcommand{\arraystretch}{1}
	\setlength{\tabcolsep}{18pt}
	\begin{tabular}{l cccc cccc}
	\toprule
	\multirow{2}{*}{Method} & \multicolumn{4}{c}{HaLo-L (\%) $\uparrow$} & \multicolumn{4}{c}{HaLo-H (\%) $\uparrow$} \\
	\cmidrule(lr){2-5} \cmidrule(lr){6-9}
	 & IoU & F1 & Pre & Recall & IoU & F1 & Pre & Recall \\
	\midrule
	LWGANet-L2 $\ast$       & 66.08 & 79.58 & 82.86 & 76.55 & 69.72 & 82.16 & 86.10 & 78.56 \\
	LWGANet-L2              & 66.81 & 80.11 & 82.68 & 77.68 & 68.84 & 81.54 & \textbf{87.93} & 76.02 \\
	HaLoBuild-Net $\ast$ & 67.53 & 80.62 & 83.70 & 77.76 & 69.95 & 82.32 & 87.17 & 77.97 \\
	HaLoBuild-Net        & \textbf{68.90} & \textbf{81.59} & \textbf{84.16} & \textbf{79.10} & \textbf{70.88} & \textbf{82.96} & 86.92 & \textbf{79.35} \\
	\bottomrule
	\end{tabular}
	\label{enhance}
\end{table*}

\subsubsection{Superiority of End-to-End Strategy}
To demonstrate the superiority of our end-to-end architecture, we benchmark HaLoBuild-Net against a cascaded paradigm (marked with $\ast$). In this comparison, models without the asterisk represent direct building extraction from raw degraded imagery without pre-processing. In contrast, the cascaded strategy utilizes restored images as input: as illustrated in Fig.~\ref{enhanced}, the HaLo-L subset is first enhanced by GSAD~\cite{hou2023global}, and the HaLo-H subset is dehazed via DEA-Net~\cite{chen2024dea}, before being fed into the segmentation networks. To ensure fairness and authenticity of the evaluation, all results for the cascaded experiments are reported as the average of three independent runs.

Quantitative results are summarized in Table~\ref{enhance}. LWGANet$\ast$ benefits from explicit dehazing because the removal of atmospheric scattering restores signal contrast and clears obscured boundaries, which effectively reduces the difficulty of feature extraction for standard backbones. However, our end-to-end HaLoBuild-Net still outperforms HaLoBuild-Net$\ast$ by 0.93\% in IoU. In sharp contrast to the hazy results, a counterintuitive performance drop is observed on the low-light HaLo-L subset after GSAD enhancement; for instance, the IoU of HaLoBuild-Net$\ast$ decreases from 68.90\% to 67.53\%. This performance degradation in the cascaded HaLo-L task reveals a decoupling between perceptual aesthetics and semantic precision. Specifically, this gap stems from three core factors:
\begin{itemize}
    \item \textbf{Generation of Artifacts:} Enhancement often hallucinates building-like patterns in dark regions, leading to false positives in the segmentation network.
    \item \textbf{Loss of Edge Sharpness:} Strong denoising during enhancement over-smooths images, erasing contrast and boundaries between buildings and backgrounds.
    \item \textbf{Distorted Feature Distribution:} Brightness amplification magnifies sensor noise, creating a statistical domain shift that confuses the segmentation head.
\end{itemize}

In contrast, HaLoBuild-Net bypasses these inherent bottlenecks by directly learning task-specific features from raw degraded observations. Instead of performing explicit pixel-level reconstruction, our framework achieves signal-level disentanglement through simultaneous optimization in the spatial and frequency domains. This allows the model to suppress meteorological interference without the risk of creating hallucinated textures or erasing critical boundaries. By shifting the objective from visual beauty to semantic integrity, HaLoBuild-Net ensures that the refined features are inherently robust to noise and distortions, providing a more reliable solution for building extraction under hazy and low-light conditions.

\subsubsection{Performance on WHU and INRIA Datasets}\label{sec:StandardResults}\mbox{}

To verify model generalizability in conventional scenarios, we conduct comparisons on WHU and INRIA datasets, with results shown in Table~\ref{table3}. On the high-resolution WHU dataset, HaLoBuild-Net achieves the best overall performance, reaching $91.88\%$ IoU and $95.77\%$ F1. On the INRIA dataset with diverse city styles and large scale variations, the model exhibits strong generalization, reaching $82.53\%$ IoU and $90.43\%$ F1. These results indicate that, although designed for hazy and low-light conditions, HaLoBuild-Net's multi-scale guidance and frequency enhancement do not degrade performance in clear scenes. Instead, they further improve boundary sharpness, structural integrity, and internal consistency, validating the general applicability of the proposed architecture.
\begin{table*}[t]
	\centering
	\scriptsize
	\caption{Comparison Results on WHU and INRIA Datasets. $\#$ Indicates that the Results are Directly Cited from the Original Publications due to the Unavailability of Publicly Released Source Codes for Reproduction. } \vspace{-2mm}
	\label{tab:new_comparison}
	\renewcommand{\arraystretch}{1.0}
	\setlength{\tabcolsep}{11pt} 
	\begin{tabular}{l c c cccc cccc}
	\toprule
	\multirow{2}{*}{Method} & \multirow{2}{*}{Publication} 
	& \multirow{2}{*}{\shortstack[c]{Params.\\ (M) $\downarrow$ }} 
	& \multicolumn{4}{c}{WHU (\%) $\uparrow$} 
	& \multicolumn{4}{c}{INRIA (\%) $\uparrow$} \\
	\cmidrule(r){4-7} \cmidrule(l){8-11} 
	& & & IoU & F1 & Pre & Recall
	& IoU & F1 & Pre & Recall \\
	\midrule
	UNetFormer~\cite{Wang_2022}     & ISPRS 2022 & 11.7 & 89.84 & 94.65 & 94.45 & 94.85 & 79.83 & 88.78 & 89.63 & 87.95 \\
	BOMSC-Net~\cite{9716137} $\#$   & TGRS 2022  & 129.3	& \textcolor{green}{90.15} & 94.80 & 95.14 & 94.50 & 78.18 & 87.75 & 87.93 & 87.58 \\
	BuildFormer~\cite{wang2022building}    & TGRS 2022  & 40.5 & 89.83 & 94.64 & 93.90 & \textcolor{green}{95.39} & 81.03 & 89.52 & 90.60 & 88.47 \\
	PolyBuilding~\cite{HU202315} $\#$ &ISPRS 2023 & 40.2	& - & - & - & 81.20 & - & - & - & - \\ 
	BCTNet~\cite{10086688} $\#$      & TGRS 2023  & 78.3 & 91.15 & \textcolor{green}{95.37} & 95.47 & 95.27 & -     & -     & -     & -     \\
	FD-Net~\cite{10155230} $\#$      & TGRS 2023  & 19.7 & 91.14 & 95.36 & 95.27 & \textcolor{blue}{95.46} & -     & -     & -     & -     \\
	Easy-Net~\cite{10376153} $\#$    & TGRS 2024  & 3.9 & 89.16 & 94.20 & 94.94 & 93.48 & -     & -     & -     & -     \\
	HD-Net~\cite{LI202451} $\#$ & ISPRS 2024 & 13.9 & 90.19 & 94.84 & 95.00 & 94.68 & \textcolor{green}{82.10}   & \textcolor{green}{90.17}     & \textcolor{green}{91.10}     & \textbf{\textcolor{red}{89.26}}     \\
	EGAFNet~\cite{10819433} $\#$     & TGRS 2025  & 44.6 & 89.30 & 94.35 & 93.82 & 94.88 & -     & -     & -     & -     \\
	LOGCAN++~\cite{ma2025logcan}     & TGRS 2025  & 31.1 & \textcolor{blue}{91.51} & \textcolor{blue}{95.57} & \textbf{\textcolor{red}{96.15}} & 94.99 & \textcolor{blue}{82.13} & \textcolor{blue}{90.19} & \textcolor{blue}{91.52} & \textcolor{green}{88.90} \\
	AFENet~\cite{10955240}         & TGRS 2025  & 20.2 & 89.36 & 94.38 & \textcolor{green}{95.68} & 93.11 & 80.28 & 89.06 & 90.91 & 87.30 \\
	UMformer~\cite{10969832}       & TGRS 2025  & 12.4 & 89.90 & 94.68 & 95.15 & 94.22 & 79.66 & 88.68 & 89.75 & 87.63 \\
	\midrule
	HaLoBuild-Net (ours) & - & 23.9 
	& \textbf{\textcolor{red}{91.88}} & \textbf{\textcolor{red}{95.77}} & \textcolor{blue}{96.03} & \textbf{\textcolor{red}{95.51}}
	& \textbf{\textcolor{red}{82.53}} & \textbf{\textcolor{red}{90.43}} & \textbf{\textcolor{red}{91.80}} & \textcolor{blue}{89.10} \\
	\bottomrule
	\end{tabular}
	\label{table3}
\end{table*}
		
	\begin{figure*}[t]
		\centering 
		\includegraphics[width=0.8\textwidth]{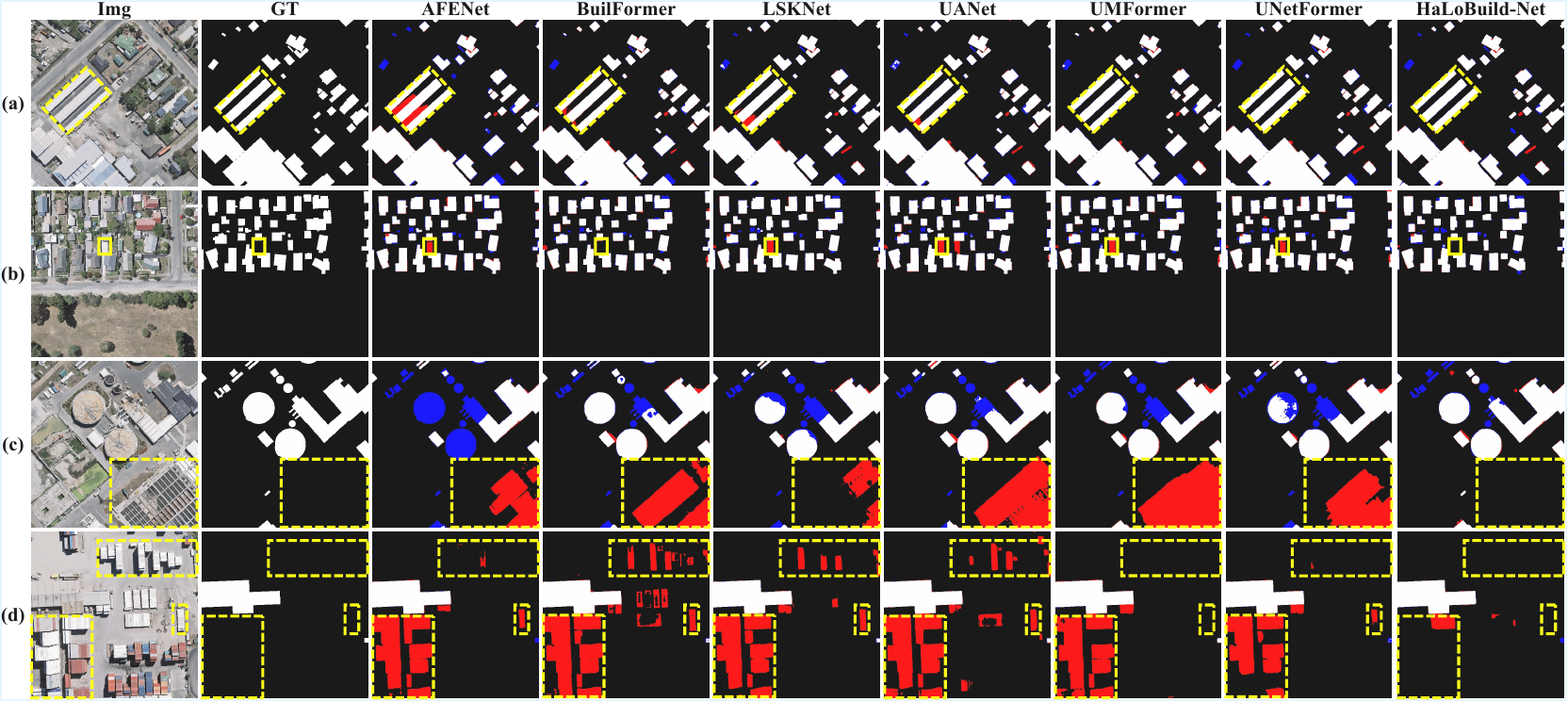}\vspace{-2mm}
		\caption{Visual comparison of building extraction results on WHU dataset. It validates the model's general applicability in conventional clear-weather scenarios. It highlights the network's proficiency in fine boundary delineation and strong discrimination capability against homomorphic interference.}
		\label{fig8}
	\end{figure*}
Fig.~\ref{fig8} shows representative visualization results on the WHU test set. Specifically, (a) and (b) highlight the model's proficiency in fine boundary delineation, effectively overcoming the adhesion effect caused by highly similar textures and precisely distinguishing buildings from adjacent cement roads. (c) and (d) demonstrate its strong semantic discrimination capability when facing homomorphic interference. Whether dealing with sewage treatment ponds exhibiting building-like geometric features or dense containers with shadows, our model penetrates surface appearances to capture deeper semantics, correctly identifying them as non-building backgrounds and effectively suppressing false detections of non-permanent facilities and regular geometric objects. Overall, HaLoBuild-Net exhibits excellent semantic consistency and interference robustness in high-resolution conventional scenes, proving its reliability in handling complex texture and geometric confusions.

\subsubsection{Performance on LoveDA Dataset}\label{sec:MultiClassResults}\mbox{}
To further validate the generalization capability of HaLoBuild-Net in complex multi-class land cover scenarios, we conducted quantitative evaluation on the highly challenging LoveDA dataset, which encompasses significant appearance differences between urban and rural areas. As shown in Table~\ref{table4}, our method achieves substantial overall superiority, with an mIoU of $54.04\%$, outperforming all comparison methods. Specifically, for the core building class, HaLoBuild-Net attains an excellent IoU of $60.25\%$, demonstrating its precise building capture ability in complex backgrounds. More notably, on road and background classes that are easily confused with buildings, our method achieves the highest accuracies of $59.81\%$ and $47.72\%$, respectively. This indicates that HaLoBuild-Net possesses strong semantic discrimination capability and maintains robust feature representation under multi-class interference.
\begin{table*}[t]
	\centering
	\scriptsize
	\caption{Comparison of IoU Results on the LoveDA Dataset. The Best, Second-best, and Third-best Results are Highlighted in \textcolor{red}{red}, \textcolor{blue}{blue}, and \textcolor{green}{green}.} \vspace{-2mm}
	\label{tab:loveda_comparison}
	\renewcommand{\arraystretch}{1.0}
	\setlength{\tabcolsep}{11pt} 
	\begin{tabular}{l c cccccccc} 
	\toprule
	\multirow{2}{*}{Method} & \multirow{2}{*}{Publication} 
	& \multicolumn{8}{c}{LoveDA (\%) $\uparrow$} \\
	\cmidrule(l){3-10} 
	& & Background & Building & Road & Water & Barren & Forest & Agricultural & mIoU \\
	\midrule
	UNetFormer~\cite{Wang_2022}     & ISPRS 2022 & 44.66 & 56.91 & 54.41 & 79.24 & 16.78 & \textcolor{green}{46.40} & 61.41 & 51.40 \\
	RSSFormer~\cite{10026298} $\#$        & TIP 2023    & \textbf{\textcolor{red}{52.38}} & \textbf{\textcolor{red}{60.71}} & 55.21 & 76.29 & 18.73 & 45.39 & 58.33 & 52.43 \\
	CMTFNet~\cite{10247595} $\#$          & TGRS 2023   & 46.20 & 58.57 & 56.06 & 79.60 & 19.88 & \textcolor{blue}{46.60} & \textcolor{blue}{63.83} & 52.96 \\
	LoveNAS~\cite{WANG2024265} $\#$       & ISPRS 2024  & 45.39 & 58.86 & \textcolor{blue}{59.45} & 79.39 & 13.76 & 43.94 & \textbf{\textcolor{red}{65.64}} & 52.34 \\
	DecoupleNet-D2~\cite{10685518} $\#$& TGRS 2024   & 45.30 & 59.50 & 56.30 & \textcolor{green}{80.60} & \textcolor{green}{20.90} & 46.20 & 63.10 & \textcolor{green}{53.10} \\
	LSKNet-S~\cite{li2025lsknet}           & IJCV 2024   & 46.44 & 58.19 & 53.86 & 78.32 & \textcolor{blue}{23.52} & 45.73 & \textcolor{green}{63.43} & 52.78 \\
	UMFormer~\cite{10969832}         & TGRS 2025   & 45.72 & 56.84 & 53.90 & 80.09 & 14.40 & 46.24 & 60.11 & 51.04 \\
	AFENet~\cite{10955240}         & TGRS 2025   & 44.99 & \textcolor{green}{60.11} & 56.16 & \textbf{\textcolor{red}{81.39}} & 20.11 & \textbf{\textcolor{red}{47.45}} & 60.14 & 52.91 \\
	LWGANet-L2~\cite{lu2025lwganet} $\#$    & AAAI 2026   & \textcolor{green}{46.76}     & 59.56     & \textcolor{green}{56.73}     & 79.57     & \textbf{\textcolor{red}{23.58}}     & 46.28     & 62.42     & \textcolor{blue}{53.60} \\
	\midrule
	\textbf{HaLoBuild-Net (ours)} & - & \textcolor{blue}{47.72} & \textcolor{blue}{60.25} & \textbf{\textcolor{red}{59.81}} & \textcolor{blue}{81.15} & 20.31 & 45.75 & 63.29 & \textbf{\textcolor{red}{54.04}} \\
	\bottomrule
	\end{tabular}
	\label{table4}
\end{table*}

\begin{table*}[htbp]
    \centering \scriptsize
    \caption{Ablation Study of the Proposed Modules on HaLo-L and HaLo-H Datasets. The First Row Represents the Baseline Model. Red and Blue Numbers in Parentheses Indicate Performance Gains ($+$) and Degradations ($-$) Relative to the Baseline, Respectively. }\vspace{-2mm}
    \renewcommand{\arraystretch}{1.1}
    \newcommand{\gain}[1]{~\makebox[3.2em][l]{\textcolor{red}{\scriptsize (+#1)}}}
    \newcommand{\loss}[1]{~\makebox[3.2em][l]{\textcolor{blue}{\scriptsize (-#1)}}}
    
    \setlength{\tabcolsep}{9pt}
    \begin{tabular}{ccc|cccc|cccc}
    \hline
    \multirow{2}{*}{SFFM} & \multirow{2}{*}{MGFM} & \multirow{2}{*}{GMGM} & \multicolumn{4}{c|}{HaLo-L (\%) $\uparrow$} & \multicolumn{4}{c}{HaLo-H (\%) $\uparrow$} \\ \cline{4-11} 
    &  &  & IoU & F1 & Pre & Recall & IoU & F1 & Pre & Recall \\ \hline
    &  &  & 66.40 & 79.81 & 84.30 & 75.78 & 67.92 & 80.90 & 86.74 & 75.79 \\
    \checkmark &  &  & 67.28 \gain{0.88} & 80.44 \gain{0.63} & 82.80 & 78.21 & 68.86 \gain{0.94} & 81.56 \gain{0.66} & 87.93 & 76.05 \\
    & \checkmark &  & 68.08 \gain{1.68} & 81.01 \gain{1.20} & 83.93 & 78.27 & 69.63 \gain{1.71} & 82.10 \gain{1.20} & 86.65 & 78.00 \\
    &  & \checkmark & 67.65 \gain{1.25} & 80.71 \gain{0.90} & 83.38 & 78.20 & 68.96 \gain{1.04} & 81.63 \gain{0.73} & 87.63 & 76.40 \\
    \checkmark & \checkmark &  & 67.69 \gain{1.29} & 80.73 \gain{0.92} & 83.52 & 78.12 & 70.38 \gain{2.46} & 82.61 \gain{1.71} & 86.81 & 78.80 \\
    \checkmark &  & \checkmark & 67.77 \gain{1.37} & 80.79 \gain{0.98} & 83.54 & 78.22 & 69.68 \gain{1.76} & 82.13 \gain{1.23} & 87.10 & 77.70 \\
    & \checkmark & \checkmark & 67.94 \gain{1.54} & 80.91 \gain{1.10} & 82.12 & \textbf{79.74} & 69.86 \gain{1.94} & 82.25 \gain{1.35} & \textbf{88.03} & 77.19 \\ 
    \checkmark & \checkmark & \checkmark & \textbf{68.90} \gain{2.50} & \textbf{81.59} \gain{1.78} & \textbf{84.16} & 79.10 & \textbf{70.88} \gain{2.96} & \textbf{82.96} \gain{2.06} & 86.92 & \textbf{79.35} \\ \hline
    \end{tabular}
    \label{table5}%
\end{table*}

\begin{figure}[t]
\centering 
\includegraphics[width=0.48\textwidth]{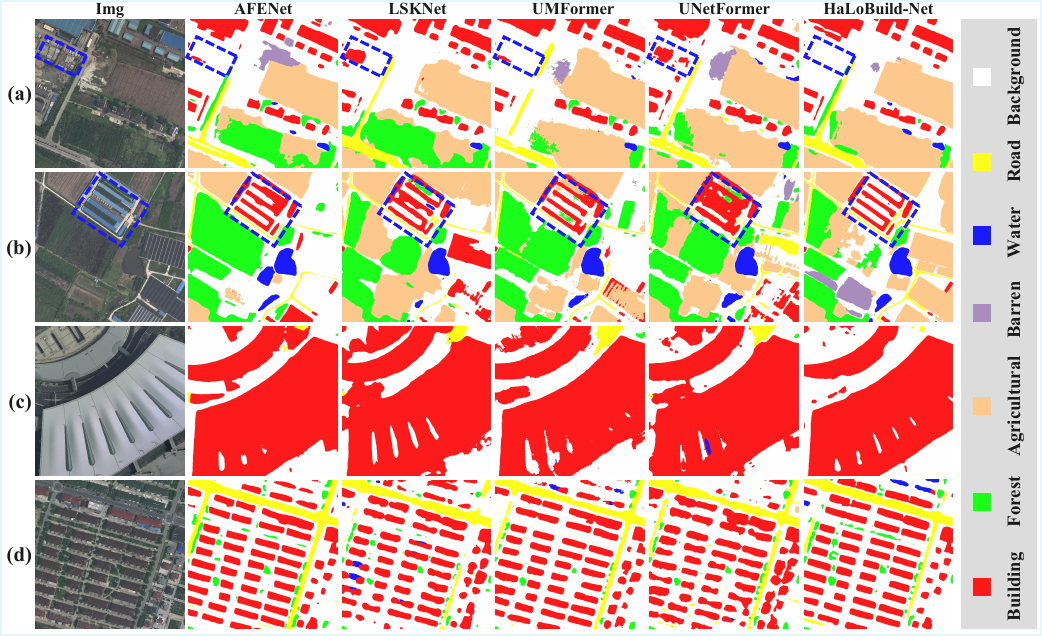} \vspace{-2mm}
\caption{Visual comparison of building extraction results on LoveDA dataset. It highlights the model's generalization capability in complex multi-class urban and rural environments. It demonstrates advantages in semantic discrimination, global context modeling, and fine boundary delineation for dense small-scale targets.}
\label{fig9}
\end{figure}
Fig.~\ref{fig9} presents visualization results on the LoveDA test set. Specifically, (a) highlights the model's advantage in semantic discrimination, effectively filtering out confusing non-building interference such as containers and avoiding misdetections caused by homomorphic ambiguity. (b) demonstrates its feature decoupling capability in complex backgrounds, precisely separating buildings from closely adjacent vegetation. (c) reflects its global context modeling ability, ensuring continuity and completeness inside large-scale buildings while avoiding fragmentation. (d) emphasizes its fine boundary delineation for dense small-scale targets, successfully preserving regular contours between buildings. 

\section{Ablation Study}

\subsection{Analysis of Individual Modules}
As summarized in Table~\ref{table5}, we first evaluate the impact of introducing each module independently. The results demonstrate that all three proposed modules---SFFM, MGFM, and GMGM---consistently yield performance improvements across both datasets. Among them, the MGFM emerges as the strongest individual contributor, achieving IoU improvements of $1.68\%$ and $1.71\%$ on HaLo-L and HaLo-H, respectively. This highlights the fundamental importance of bidirectional semantic-spatial calibration in resolving feature contamination and boundary diffusion under adverse weather. The GMGM provides significant gains of $1.25\%$ and $1.04\%$, validating that global topological anchoring is essential for compensating for the visibility loss and structural ambiguity in degraded imagery. Meanwhile, the SFFM independently achieves improvements of $0.88\%$ and $0.94\%$. These positive gains empirically support our frequency-domain modeling strategy: even as a standalone component, leveraging stable low-frequency structural anchors effectively recalibrates feature responses against meteorological interference and spectral distortions.
        
\subsection{Module Synergy and Complementarity}
Subsequent combination experiments reveal critical synergies that validate the closed-loop architecture of HaLoBuild-Net. In hazy scenarios, while SFFM alone provides a moderate gain, its integration with MGFM triggers a more substantial performance leap, reaching a $2.46\%$ IoU improvement. This confirms that the bidirectional calibration from MGFM provides the necessary semantic context, which enables SFFM to more accurately distinguish stable building topologies from haze-induced spectral artifacts in the Fourier domain. In low-light scenarios, the combination of SFFM and GMGM results in a $1.37\%$ IoU boost, confirming that in extreme darkness, recovering fine topological details requires the joint optimization of global positional priors and signal-level frequency recalibration. 
        
Ultimately, the complete HaLoBuild-Net achieves the best performance on both datasets, reaching peak IoU scores of 68.90\% ($+2.50\%$) on HaLo-L and 70.88\% ($+2.96\%$) on HaLo-H. These results fully validate that the proposed framework effectively synergizes frequency-aware modulation, topological anchoring, and semantic calibration, enabling robust building extraction across diverse and complex adverse conditions.

\begin{table*}[htbp]
\centering \scriptsize
\caption{Performance Comparison of Different Backbones on HaLo-L and HaLo-H Datasets. The best results are highlighted in \textbf{bold}.} \vspace{-2mm}
\renewcommand{\arraystretch}{1}
\setlength{\tabcolsep}{14pt}
\begin{tabular}{l c cccc cccc}
\toprule
\multirow{2}{*}{Backbone} & \multirow{2}{*}{\shortstack[c]{Params.\\ (M) $\downarrow$}}
& \multicolumn{4}{c}{HaLo-L (\%) $\uparrow$} & \multicolumn{4}{c}{HaLo-H (\%) $\uparrow$} \\
\cmidrule(r){3-6} \cmidrule(l){7-10}
& & Iou & F1 & Pre & Recall & Iou & F1 & Pre & Recall \\
\midrule
ResNet-18 & 11.7 & 59.36 & 74.50 & 77.29 & 71.90 & 65.81 & 79.38 & 82.75 & 76.28 \\
ResNet-34 & 21.8 & 67.90 & 80.88 & 83.05 & 78.82 & 65.65 & 79.26 & 82.10 & 76.61 \\
MobileNetV2-1.4 & 6.9 & 66.81 & 80.10 & \textbf{84.26} & 76.33 & 67.35 & 80.49 & 81.80 & 79.22 \\
LWGANet-L2 (ours) & 13.0 & \textbf{68.90} & \textbf{81.59} & 84.16 & \textbf{79.10} & \textbf{70.88} & \textbf{82.96} & \textbf{86.92} & \textbf{79.35} \\
\bottomrule
\end{tabular}
\label{table6}
\end{table*}

\begin{table*}[htbp]	\centering \scriptsize 
    \caption{Performance Comparison of Spatial and Frequency Paths within the SFFM. The best results are highlighted in \textbf{bold}.}\vspace{-2mm}
        \renewcommand{\arraystretch}{1}
        \setlength{\tabcolsep}{18pt}
        \begin{tabular}{l cccc cccc}
        \toprule 
        \multirow{2}{*}{Method} & \multicolumn{4}{c}{HaLo-L (\%) $\uparrow$} & \multicolumn{4}{c}{HaLo-H (\%) $\uparrow$} \\
        \cmidrule(r){2-5} \cmidrule(l){6-9}
        & IoU & F1 & Pre & Recall & IoU & F1 & Pre & Recall \\
        \midrule
        SFFM\_sp  & 68.24 & 81.12 & 83.52 & 78.85 & 68.59 & 81.37 & 84.06 & 78.85 \\
        SFFM\_fre & 67.84 & 80.84 & 83.92 & 77.98 & 70.03 & 82.37 & \textbf{87.04} & 78.18 \\
        SFFM (ours) & \textbf{68.90} & \textbf{81.59} & \textbf{84.16} & \textbf{79.10} & \textbf{70.88} & \textbf{82.96} & 86.92 & \textbf{79.35} \\
        \bottomrule
        \end{tabular}
        \label{table_sffm_sp_fre}
        \end{table*}
    
    \begin{table*}[htbp]
        \centering \scriptsize 
        \caption{Ablation Study on the Low-Frequency Cropping Ratio within the SFFM. The Ratio Determines the Size of the Central Region Extracted from the FFT Spectrum, with Results Evaluated on both HaLo-L and HaLo-H Datasets. The best results are highlighted in \textbf{bold}.} \vspace{-2mm}
        \renewcommand{\arraystretch}{1}
        \setlength{\tabcolsep}{17pt}
        \begin{tabular}{l cccc cccc}
        \toprule
        \multirow{2}{*}{Cropping Ratio} & \multicolumn{4}{c}{HaLo-L (\%) $\uparrow$} & \multicolumn{4}{c}{HaLo-H (\%) $\uparrow$} \\
        \cmidrule(r){2-5} \cmidrule(l){6-9}
        & IoU & F1 & Pre & Recall & IoU & F1 & Pre & Recall \\
        \midrule
        $H \times W$       & 68.13 & 81.05 & 81.99 & 80.12 & 70.19 & 82.49 & 86.99 & 78.42 \\
        $H/2 \times W/2$   & 66.15 & 79.63 & \textbf{87.42} & 73.11 & 70.43 & 82.65 & \textbf{87.14} & 78.60 \\
        $H/4 \times W/4$ (ours)   & \textbf{68.90} & \textbf{81.59} & 84.16 & 79.10 & \textbf{70.88} & \textbf{82.96} & 86.92 & \textbf{79.35} \\
        $H/8 \times W/8$   & 68.19 & 81.09 & 78.02 & \textbf{82.17} & 70.06 & 82.39 & 86.89 & 78.34 \\
        \bottomrule
        \end{tabular}
        \label{table7}
    \end{table*}

\subsection{Backbone Architecture Analysis}
To justify the selection of LWGANet-L2, we benchmark it against mainstream backbones including ResNet-18, ResNet-34, and MobileNetV2-1.4, with results detailed in Table~\ref{table6}. While MobileNetV2-1.4 offers the lowest parameter count (6.9 M), its feature extraction capability proves insufficient for hazy and low-light conditions, resulting in suboptimal IoU scores. Conversely, although ResNet-34 achieves competitive performance, it incurs a heavier computational burden (21.8 M). Our selected LWGANet-L2 strikes the optimal balance between efficiency and accuracy. With parameters (13.0 M) comparable to ResNet-18, it outperforms ResNet-18 by 9.54\% and 5.07\% in IoU on HaLo-L and HaLo-H, respectively. Furthermore, it surpasses the heavier ResNet-34, demonstrating that the Lightweight Group Attention (LWGA) mechanism is particularly effective at capturing multi-scale building features under low-contrast and hazy conditions.

\subsection{Ablation Study of SFFM}
We evaluate the internal mechanisms of SFFM by decoupling it into the spatial path and frequency path. As shown in Table~\ref{table_sffm_sp_fre}, SFFM\_sp performs better on HaLo-L (68.24\% IoU) as spatial attention effectively recovers structures in low light. Conversely, SFFM\_fre excels on HaLo-H (70.03\% IoU) by leveraging the stable low-frequency spectrum to mitigate the impact of atmospheric scattering. The complete SFFM achieves the best results on both datasets, proving the necessity of collaborative dual-domain modulation.
    
We also analyze the sensitivity of the cropping ratio that determines the range of frequency-domain structural anchoring. As presented in Table~\ref{table7}, utilizing the full spectrum ($H \times W$) introduces unreliable high-frequency components corrupted by weather interference, while intermediate ($H/2 \times W/2$) and overly small ($H/8 \times W/8$) ratios either fail to provide a robust global anchor or cause a loss of essential topological details. The $H/4 \times W/4$ configuration achieves the optimal performance by effectively balancing the suppression of spectral artifacts and the preservation of global building structures.

\section{Conclusion and Future Work}
In this work, we addressed the persistent challenge of building extraction from optical remote sensing imagery under hazy and low-light conditions. To resolve the data scarcity, we established HaLoBuilding, the first optical benchmark specifically curated for hazy and low-light environments, utilizing a same-scene multitemporal pairing strategy to ensure high-fidelity, pixel-level label alignment. Furthermore, we proposed HaLoBuild-Net, a novel end-to-end framework that achieves direct building extraction without explicit image enhancement. The network facilitates dual-domain collaborative optimization via the Spatial-Frequency Focus Module (SFFM). Moreover, the integration of the Global Multi-scale Guidance Module (GMGM) and the Mutual-Guided Fusion Module (MGFM) anchors building topologies and bridges semantic gaps through bidirectional semantic-spatial calibration, respectively. Extensive experiments confirm that HaLoBuild-Net not only achieves state-of-the-art performance on the HaLoBuilding benchmark but also significantly outperforms conventional cascaded paradigms. Furthermore, our model demonstrates robust generalization across conventional datasets including WHU, INRIA, and LoveDA. Future work will focus on improving model efficiency for real-time edge deployment and extending the framework to multi-class semantic segmentation under a broader range of extreme weather scenarios, aiming to build a robust foundation model for all-weather remote sensing perception. This ultimately contributes to advancing reliable and automated remote sensing analysis for diverse real-world applications.
\bibliographystyle{IEEEtran} 
\bibliography{b_refs}       

\vfill
\end{document}